\documentclass[acmtog,table,prologue,dvipsnames]{acmart} 

\makeatletter

\usepackage{booktabs} 

\usepackage[ruled]{algorithm2e} 

\SetAlFnt{\small}
\SetAlCapFnt{\small}
\SetAlCapNameFnt{\small}
\SetAlCapHSkip{0pt}

\newcommand{\etal}{\textit{et al}. }

\usepackage{amsmath}
\usepackage{color}
\usepackage{booktabs}
\usepackage{caption}
\usepackage{subcaption}
\usepackage{enumitem}
\usepackage{multirow}
\definecolor{darkorange}{rgb}{1.0, 0.55, 0.0}

\newcommand{\change}[1]{#1}

\newcommand{\apref}[1]{Appx.~\ref*{#1}}

\newcommand{\cellfirst}{\cellcolor{Red!40}}
\newcommand{\cellsecond}{\cellcolor{Orange!25}}
\newcommand{\cellthird}{\cellcolor{Yellow!25}}

\usepackage{amsmath}

\acmJournal{TOG}

\setcopyright{acmcopyright}

\begin{document}
\title{Promptable Game Models: Text-Guided Game Simulation via Masked Diffusion Models}
%

\author{Willi Menapace}
\authornote{Work performed while the author was an intern at Snap Inc.}
\affiliation{
  \institution{University of Trento}
  \country{Italy}
}
\email{willi.menapace@unitn.it}
\author{Aliaksandr Siarohin}
\affiliation{
  \institution{Snap Inc.}
  \country{USA}
}
\email{asiarohin@snapchat.com}
\author{St\'{e}phane Lathuili\`{e}re}
\affiliation{
  \institution{LTCI, T\'{e}l\'{e}com Paris, Institut Polytechnique de Paris}
  \country{France}
}
\email{stephane.lathuiliere@telecom-paris.fr}
\author{Panos Achlioptas}
\affiliation{
  \institution{Snap Inc.}
  \country{USA}
}
\email{pachlioptas@gmail.com}
\author{Vladislav Golyanik}
\affiliation{
  \institution{MPI for Informatics, SIC}
  \country{Germany}
}
\email{golyanik@mpi-inf.mpg.de}
\author{Sergey Tulyakov}
\affiliation{
  \institution{Snap Inc.}
  \country{USA}
}
\email{stulyakov@snapchat.com}
\author{Elisa Ricci}
\affiliation{
  \institution{University of Trento, Fondazione Bruno Kessler}
  \country{Italy}
}
\email{e.ricci@unitn.it}

\renewcommand\shortauthors{Menapace, W. et al}

\begin{abstract}

\change{Neural video game simulators emerged as powerful tools to generate and edit videos. Their idea is to represent games as the evolution of an environment's state driven by the actions of its agents. While such a paradigm enables users to \emph{play} a game action-by-action, its rigidity precludes more semantic forms of control. To overcome this limitation, we augment game models with \emph{prompts} specified as a set of \emph{natural language} actions and \emph{desired states}.} The result---a \change{Promptable Game Model (PGM)}---makes it possible for a user to \emph{play} the game by prompting it with high- and low-level action sequences. Most captivatingly, our \change{PGM} unlocks the \emph{director's mode}, where the game is played by  specifying goals for the agents in the form of \change{a prompt}.
This requires learning ``game AI'', encapsulated by our animation model, to navigate the scene using high-level constraints, play against an adversary, and devise a strategy to win a point. To render the resulting state, we use a compositional NeRF representation encapsulated in our synthesis model. To foster future research, we present newly collected, annotated and calibrated Tennis and Minecraft datasets. Our method significantly outperforms existing neural video game simulators in terms of rendering quality and unlocks applications beyond the capabilities of the current state of the art. Our framework, data, and models are available at \href{https://snap-research.github.io/promptable-game-models/}{\url{snap-research.github.io/promptable-game-models}}.

\end{abstract}

\setcopyright{acmlicensed}
\acmJournal{TOG}
\acmYear{2023} \acmVolume{1} \acmNumber{1} \acmArticle{1} \acmMonth{1} \acmPrice{15.00}\acmDOI{10.1145/3635705}

%
%


%
%


\begin{teaserfigure}
\centering
\vspace{-2mm}
  \includegraphics[width=\textwidth]{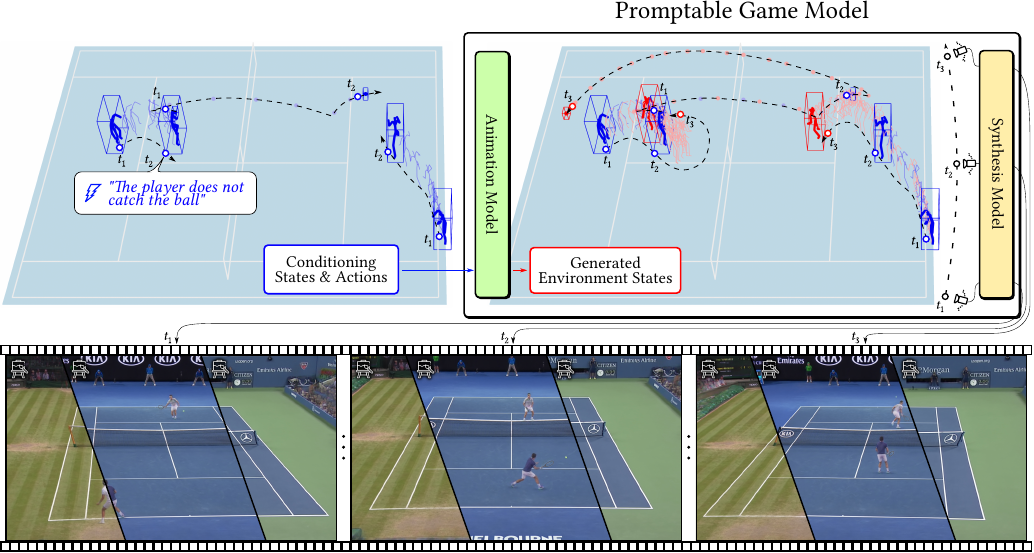}
  \caption{
  We propose \change{Promptable Game Models (PGMs), controllable models of games} that are learned from annotated videos. Our \change{PGM} enables the generation of videos using \change{prompts}, a wide spectrum of \textcolor{blue}{conditioning signals} such as player poses, object locations, and detailed textual actions (see \includegraphics[]{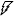}) indicating what each player should do. Our \emph{Animation Model} uses this information to generate future, past, or interpolated \textcolor{red}{environment states} according to the learned game dynamics. At this stage, the model is able to perform complex action reasoning such as generating a winning shot if the action ``the [other] player does not catch the ball'' is specified, as shown in the figure.
  To accomplish this goal, the model decides that the bottom player should hit the ball with a ``lob'' shot, sending the ball high above the opponent, who is unable to catch it.
  Our model renders the scene from a user-defined viewpoint (see \includegraphics[]{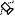}) using a \emph{Synthesis Model} where the style of the scene (see \includegraphics[]{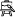}) can be controlled explicitly.
  }
  \label{fig:teaser}
\end{teaserfigure}

\maketitle

\newcommand{\netstyle}{\mathcal{E}}
\newcommand{\netenhancer}{\mathcal{F}}
\newcommand{\netcanonical}{\mathcal{C}}
\newcommand{\netdeformation}{\mathcal{D}}
\newcommand{\netanimation}{\mathcal{A}}
\newcommand{\nettext}{\mathcal{T}}
\newcommand{\nettextenc}{\mathcal{T}_\mathrm{enc}}
\newcommand{\nettextagg}{\mathcal{T}_\mathrm{agg}}
\newcommand{\netprojection}{\mathcal{P}}
\newcommand{\settext}{\mathbb{L}}
\newcommand{\setstate}{\mathbb{S}}


\newcommand{\vecbbox}{\mathbf{b}}
\newcommand{\vecbboxtwo}{\mathbf{b}^\mathrm{2D}}
\newcommand{\vecbboxthree}{\mathbf{b}^\mathrm{3D}}

\newcommand{\vecpoint}{\mathbf{x}}
\newcommand{\vecpointbbox}{\mathbf{x}_b}
\newcommand{\vecpointcanon}{\mathbf{x}_c}
\newcommand{\vecstyle}{\boldsymbol{\omega}}
\newcommand{\vecpose}{\boldsymbol{\pi}}
\newcommand{\vecdensity}{\mathbf{\sigma}}
\newcommand{\veccolor}{\mathbf{c}}
\newcommand{\vecdirection}{\mathbf{d}}
\newcommand{\vecfieldfeature}{\mathbf{f}}
\newcommand{\vecblendweights}{\mathbf{w}}
\newcommand{\vectranslation}{\mathbf{tr}}
\newcommand{\vecdiffnoise}{\boldsymbol{\epsilon}}
\newcommand{\vecdiffnoisepred}{\boldsymbol{\epsilon}^p}
\newcommand{\vecdiffalpha}{\alpha}
\newcommand{\vecdiffbeta}{\beta}
\newcommand{\vecdiffdata}{\mathbf{x}}
\newcommand{\vecactionemb}{\mathbf{a^\mathrm{emb}}}
\newcommand{\vecsequence}{\mathbf{s}}
\newcommand{\vecsequencecond}{\mathbf{s}^c}
\newcommand{\vecsequencepred}{\mathbf{s}^p}
\newcommand{\vecsequenceprednoise}{\mathbf{{s}}^p}
\newcommand{\vecsequenceemb}{\mathbf{e}}
\newcommand{\vecmask}{\mathbf{m}^\mathbf{s}}
\newcommand{\vecmaskaction}{\mathbf{m}^\mathbf{a}}
\newcommand{\vecvelocity}{\mathbf{v}}
\newcommand{\vectextaction}{\mathbf{a}}
\newcommand{\vectextactioncond}{\mathbf{a}^c}


\newcommand{\tensimage}{\mathbf{I}}
\newcommand{\tensimagerec}{\mathbf{\hat{I}}}
\newcommand{\tensimagergb}{\mathbf{\Tilde{I}}}
\newcommand{\tensnerffeatures}{\mathbf{G}}
\newcommand{\tensvoxel}{\mathbf{V}}
\newcommand{\tensplane}{\mathbf{P}}
\newcommand{\tensvoxelsmall}{\mathbf{V}'}
\newcommand{\tensplanesmall}{\mathbf{P}'}
\newcommand{\tensblendweights}{\mathbf{W}}
\newcommand{\tensblendweightssmall}{\mathbf{W}'}
\newcommand{\tensrotation}{\mathbf{R}}


\newcommand{\numseqtimesteps}{T}
\newcommand{\numdifftimesteps}{K}
\newcommand{\numobjects}{O}
\newcommand{\numproperties}{P}
\newcommand{\numactions}{A}
\newcommand{\numactionembsize}{N_t}
\newcommand{\numembsize}{E}
\newcommand{\numpropertiesembsize}{N_d}
\newcommand{\numjoints}{J}
\newcommand{\numfieldfeat}{F}
\newcommand{\numfieldfeatsmall}{F'}
\newcommand{\numimgheight}{H}
\newcommand{\numimgwidth}{W}
\newcommand{\numvoxelheight}{H_{V}}
\newcommand{\numvoxelwidth}{W_{V}}
\newcommand{\numvoxeldepth}{D_{V}}
\newcommand{\numvoxelheightsmall}{H'_{V}}
\newcommand{\numvoxelwidthsmall}{W'_{V}}
\newcommand{\numvoxeldepthsmall}{D'_{V}}
\newcommand{\numplaneheight}{H_{P}}
\newcommand{\numplanewidth}{W_{P}}
\newcommand{\numplaneheightsmall}{H'_{P}}
\newcommand{\numplanewidthsmall}{W'_{P}}
\newcommand{\numblendweightsheight}{H_{W}}
\newcommand{\numblendweightswidth}{W_{W}}
\newcommand{\numblendweightsdepth}{D_{W}}
\newcommand{\numblendweightsheightsmall}{H'_{W}}
\newcommand{\numblendweightswidthsmall}{W'_{W}}
\newcommand{\numblendweightsdepthsmall}{D'_{W}}


\newcommand{\ray}{r}
\newcommand{\difftimestep}{k}
\newcommand{\seqtimestep}{t}
\newcommand{\framerate}{\nu}
\newcommand{\airdrag}{C}

\section{Introduction}
\label{sec:introduction}

\change{Recent video generation methods, thanks to their training on extensive web-scale datasets \cite{schuhmann2022laionb}, exhibit a remarkable capacity for generating a vast amount of different concepts and scenes \cite{ho2022imagenvideo,singer2022makeavideo,blattmann2023videoldm}. Despite this, their generic nature hinders their comprehension of the dynamics of the modeled scenes. When generating or editing videos of a game, such as a tennis match, this limitation impedes their ability to attain precise control of the player movements or to devise optimal strategies to reach desired states of the game, such as victory over the opponent.}

Neural video game simulators, \change{a growing category of video generation methods, make an important step in this direction by focusing on modeling the dynamics of an environment, often a sports or computer game, with high fidelity and degree of control, and} show that annotated videos can be used to learn to generate videos interactively \cite{Kim2020_GameGan,kim2021drivegan,menapace2021pvg,huang2022layered,davtyan2022glass} and build 3D environments where agents can be controlled through a set of discrete actions \cite{Menapace2022PlayableEnvironments}. However, when applied to complex or real-world environments, these works present several limitations: they do not accurately model the game's dynamics, do not model physical interactions of objects in 3D space, do not learn precise controls, do not allow for high-level goal-driven control of the game flow, and, finally, do not model intelligent behavior of the agents, a capability often referred to as ``game AI''.

\change{In this work, we overcome these limitations by introducing game models trained on a set of annotated videos that support complex prompts. Due to the versatility of the applications enabled by diverse prompting methods (see Sec.~\ref{sec:applications}), we call them Promptable Game Models (PGMs).} More formally, we define \change{PGMs} as those models supporting a core set of game modeling and prompting functions including rendering from a controllable viewpoint, modeling of game's dynamics, precise character control, high-level goal-driven control of the game, and game AI. 
Making a first step towards the realization of such models, we propose a framework that supports these characteristics. 

To overcome the limitations of \cite{Kim2020_GameGan,kim2021drivegan,huang2022layered,davtyan2022glass,menapace2021pvg, Menapace2022PlayableEnvironments}, 
not only we model the \emph{states} of an environment, but we also consider detailed textual representations of the actions taking place in it.  We argue that training on user commentaries describing detailed actions of a game greatly facilitates learning the dynamics of the game and game AI---important parts of \change{PGMs}---and that such commentaries are a key component in enabling a series of important model capabilities related to precise character control and high-level goal-driven control of the game flow.

In its simplest form, for games like tennis, this enables controlling each player in a precise manner with instructions such as \emph{``hit the ball with a backhand and send it to the right service box''}.

Moreover, language enables users to take the \emph{director's mode} and prompt the model with high-level game-specific scenarios or scripts, specified by means of \emph{natural language} and \emph{desired states of the environment}. As an example, given desired starting and ending states, our \change{promptable game model} can devise in-between scenarios that led to the observed outcome. Most interestingly, as shown in Fig.~\ref{fig:teaser}, given the initial states of a real tennis video in which a player lost a point, our model prompted by the command \emph{“the [other] player does not catch the ball”} can perform the necessary action to win the point.

Broadly speaking, a game maintains states of its environments \cite{Stanton2016,starke2019neural,Curtis2022}, renders them using a controllable camera, and evolves them according to user commands, actions of non-playable characters controlled by the game AI, and the game's dynamics. 
Our framework follows this high-level structure highlighted in Fig.~\ref{fig:teaser}. Our synthesis model maintains a state for every object and agent included in the game and renders them in the image space using the compositional NeRF of~\cite{Menapace2022PlayableEnvironments} followed by a learnable enhancer for superior rendering quality. 
To model the dynamics of games and game AI that determine the evolution of the environment states, we introduce an animation model. Specifically, inspired by~\cite{han2022show}, we train a \emph{non-autoregressive} \emph{text-conditioned} diffusion model which leverages masked sequence modeling \change{to express the conditioning signals corresponding to a prompt}. 
In particular, we show that using text labels describing actions happening in a game is instrumental in learning such capabilities.  While certain prior work~\cite{Kim2020_GameGan,kim2021drivegan,menapace2021pvg,Menapace2022PlayableEnvironments} explored maintaining and rendering states of games, we are not aware of any generative method that attempts to enable precise control, modeling sophisticated goal-driven game dynamics, and learning game AI to the extent explored in this paper.

The task of playing games and manipulating videos in the \emph{director's mode} has not been previously introduced in the literature. With this work, we attempt to introduce the task and set up a solid framework for future research. To do that, we collected two monocular video datasets. The first one is the Minecraft dataset containing 1.2 hours of videos, depicting a player moving in a complex environment.  
The second is a large-scale real-world dataset with 15.5 hours of high-resolution professional tennis matches. For each frame in these datasets, we provide accurate camera calibration, 3D player poses, ball localization and, most importantly, diverse and rich text descriptions of the actions performed by each player in each frame. 

In summary, our work brings the following contributions:
\begin{itemize}
\item A framework for the creation of \change{Promptable Game Models}.
It supports detailed offline rendering of high-resolution, high-frame rate videos of scenes with articulated objects from controllable viewpoints. It can generate actions specified by detailed text prompts, model opponents, and perform goal-driven generation of complex action sequences. As far as we are aware, no existing work provides this set of capabilities under comparable data assumptions.
\item A synthesis model, based on a compositional NeRF \change{backed by an efficient plane- and voxel-based object representation that operates without upsampling.} 
\change{With respect to the upsampler-based approach of \cite{Menapace2022PlayableEnvironments}, it doubles the output resolution, can synthesize small objects and does not present checkerboard upsampling artifacts.}
\item An animation model, based on a text-conditioned diffusion model with a masked training procedure, which is key to supporting complex game dynamics, object interactions, game AI, and understanding detailed actions. It unlocks applications currently out of reach of state-of-the-art neural video game simulators (see Sec.~\ref{sec:applications}). 
\item A large-scale 15h Tennis and a 1h Minecraft video datasets with camera calibration, 3D player poses, 3D ball localization, and detailed text captions.
\end{itemize}

\section{Related Work} 

Our \change{Promptable Game Model} relates to neural game simulation literature, game engines, character animation, neural rendering, sequential data generation, and text-based generation.
We review the most recent related works in this section.

\subsection{Neural video game simulation}
\label{sec:neural_game_simulation}
In the last few years, video game simulation using deep neural networks has emerged as a new research trend ~\cite{Kim2020_GameGan,kim2021drivegan,menapace2021pvg,Menapace2022PlayableEnvironments,davtyan2022glass,huang2022layered}.  \change{The objective is to train a neural network to synthesize videos based on a specific type of prompt: a sequence of actions provided at every time step.}

This problem was first addressed using training videos annotated with the corresponding action labels at each time step~\cite{oh2015action,chiappa2017recurrent,Kim2020_GameGan}. They consider a discrete action representation that is difficult to define a priori for real-world environments. More recently, \cite{kim2021drivegan} proposed a framework that uses a continuous action representation to model real-world driving scenarios. Devising a good continuous action representation for an environment, however, is complex.
To avoid this complexity, \cite{menapace2021pvg, Menapace2022PlayableEnvironments} propose to learn a discrete action representation. \cite{huang2022layered} expands on this idea by modeling actions as a learned set of geometric transformations, while \cite{davtyan2022glass} represents actions by separating them into a global shift component and a local discrete action component.

\change{Differently from our PGM}, previous works perform generation in an autoregressive manner, conditioned on the actions and, therefore, are unable to answer prompts entailing constraint- or goal-driven generation for which non-sequential conditioning is necessary. We find the proposed text-based action representation and masked training procedure to be crucial to unlocking such applications.

Among these works, \emph{Playable environments}~\cite{Menapace2022PlayableEnvironments} is the most closely related to ours. Rather than employing a 2D model, they use a NeRF-based renderer~\cite{mildenhall2020nerf} that enables them to represent complex 3D scenes. \change{We follow this high-level design but introduce a more efficient plane- and voxel-based NeRF representation that enables the rendering of outputs at double the original resolution without the use of upsampling modules which we found to be the cause of checkerboard artifacts, failures in rendering of small objects and to be prone to failure when training at higher resolutions.}
\change{In addition,} the employed discrete action representation shows limitations in complex scenarios such as tennis, where it is only able to capture the main movement directions of the players and does not model actions such as ball hitting. In contrast, we employ a text action representation that specifies actions at a fine level of granularity (i.e. which particular ball-hitting action is being performed and where the ball is sent), while remaining interpretable and intuitive for the user. \change{Lastly, we replace the adversarially-trained LSTM animation module with a more capable masked diffusion transformer.}

\subsection{Game Engines}
\label{sec:game_engines}
Game engines brought a revolution to game development by providing extensible and reusable software that can be employed to create a wide range of game models \cite{gregory2018gameengine}. Nowadays, a range of game engines exists (\emph{Unity}, \emph{Unreal}, \emph{id Tech}, \emph{Source}, \emph{CRYENGINE}, \emph{Frostbite}, \emph{RAGE}) and have grown to become vast software ecosystems. Modern game engines are organized into components including a rendering engine \cite{muller2020ncv}, a resource manager, a module for physics and collision, an animation manager and, importantly, a gameplay foundation system that models the game rules and encapsulates game AI functionalities \cite{gregory2018gameengine}. \change{The presence of these components, coupled with the labor} of a range of trained experts including software engineers, artists (animators, 3D modelers, texture and lighting artists) and game developers, \change{enables the construction of sophisticated game models supporting low-level character control and scripted agent behavior. We show that monocular videos annotated with a fraction of the effort (see \apref{ap:cost_quality}) can be used to learn models of games that support answering challenging prompts related to agent intelligence, a capability difficult to achieve through scripted agent behavior.}

\subsection{Character Animation}
\label{sec:character_animation}

Character animation is a long-standing problem in computer graphics. 
Several recent methods have been proposed that produce high-quality animations. Holden \etal \cite{holden2020learned} propose a learnable version of Motion Matching \cite{buttner2019motion} that formulates character animation as retrieval of the closest motion from a motion database and supports interaction with other characters or objects.
Other approaches model the evolution of characters using time series models conditioned on the preceding state and control signals \cite{starke2019neural,starke2020local,lee2018interactive,holden2017phase}. Starke \etal \cite{starke2019neural} propose a model based on a mixture of experts that controls character locomotion and object interactions, in a follow-up work \cite{starke2020local} they introduce local motion phases to model complex character motions and interaction with a second character.

To produce high-quality animations the methods rely on difficult-to-acquire motion capture data enriched with contact information \cite{holden2020learned,starke2019neural,starke2020local}, motion phases \cite{starke2019neural} or engineered action labels \cite{starke2019neural,starke2020local}. Additionally, handcrafted dataset-specific feature representations and mappings from user controls to such representations are often leveraged, and additional knowledge is injected through postprocessing steps such as inverse kinematics or external physics models. While these assumptions promote high-quality outputs, they come at a significant effort. In contrast, our method sidesteps these requirements by not using motion capture and basing user control on natural language that is cheaper to acquire (see \apref{ap:cost_quality}) and does not require manual engineering. Finally, character animation methods support limited goal-driven control such as interacting with a specific object while avoiding collisions \cite{starke2019neural}. In contrast, our method models complex game AI tasks such as modeling strategies to defeat the opponent, which are instrumental \change{in answering complex user prompts.}

 \subsection{Neural Rendering}
 \label{sec:neural_rendering}

Neural rendering was recently revolutionized by the advent of NeRF \cite{mildenhall2020nerf}. Several modifications of the NeRF framework were proposed to model deformable objects \cite{tretschk2021nonrigid,park2021nerfies,weng2022humannerf,park2021hypernerf,li2022tava}, and decomposed scene representations \cite{Menapace2022PlayableEnvironments,niemeyer2021giraffe,mueller2022autorf,Ost_2021_CVPR,kundu2022panoptic}. In addition, several works improved the efficiency of the original MLP representation of the radiance field \cite{mildenhall2020nerf} by employing octrees \cite{yu2021plenoctrees,martel2021acorn}, voxel grids \cite{fridovich2022plenoxels}, triplanes \cite{chan2022eg3d}, hash tables \cite{mueller2022instant}, or factorized representations \cite{chen2022tensorf}.

Our framework is most related to that of \cite{weng2022humannerf}, since we model player deformations using an articulated 3D prior and linear blend skinning (LBS) \cite{lewis2000pose}. Differently from them, however, we consider scenes with multiple players and apply our method to articulated objects with varied structures for their kinematic trees. 
While similar to the rendering framework of \cite{Menapace2022PlayableEnvironments}, our framework does not adopt computationally-inefficient MLP representations, using voxel \cite{fridovich2022plenoxels} or plane representations instead, \change{thus does not rely on upsampler networks.}

\subsection{Sequential data generation with diffusion models}
In prior work, sequential data generation was mainly addressed with auto-regressive formulations combined with adversarial~\cite{kwon2019predicting} or variational~\cite{fortuin2020gp,babaeizadeh2018stochastic} generative models. Recently, diffusion models have emerged as a promising solution to this problem leading to impressive results in multiple applications such as audio~\cite{kong2020diffwave,lam2022bddm,chen2021wavegrad,leng2022binauralgrad} and video synthesis~\cite{ho2022video,ho2022imagenvideo,singer2022makeavideo,blattmann2023videoldm}, language modeling~\cite{dieleman2022cdcd}, and human motion synthesis \cite{zhang2022motiondiffuse,dabral2022mofusion}.
Following this methodological direction~\cite{tashiro2021csdi}, introduces a score-based diffusion model for imputing missing values in time series. They introduce a training procedure based on masks that simulate missing data. This approach motivates our choice of a similar masking strategy to model the conditions entailed by the given prompt and generate the unknown environment states. In this work, we show that mask-based training is highly effective in modeling geometric properties together with textual data modalities.

\subsection{Text-based generation}
In recent years, we have witnessed the emergence of works on the problem of text-based generation.
Several works address the problem of generating images \cite{saharia2022imagen,rombach2021highresolution,ramesh2021zeroshot,ramesh2022hierarchical} and videos with arbitrary content~\cite{ho2022video,ho2022imagenvideo,singer2022makeavideo,hong2022cogvideo}, and arbitrary 3D shapes \cite{jain2021dreamfields,lin2022magic3d,achlioptas2023shapetalk}.

Han \emph{et al.} \cite{han2022show} introduced a video generation framework that can incorporate various conditioning modalities in addition to text, such as segmentation masks or partially occluded images. Their approach employs a frozen RoBERTa \cite{liu2020roberta} language model and a sequence masking technique. Fu \emph{et al.} \cite{fu2022tellmewhathappened} propose an analogous framework. Our animation framework employs a similar masking strategy, but we model text conditioning at each timestep in the sequence, use diffusion models which operate on continuous rather than discrete data, and generate scenes that can be rendered from arbitrary viewpoints. 

More relevant to our work, several papers introduced models to generate human motion sequences from text \cite{guy2022motionclip,TEACH}. 
Recently, diffusion models have shown strong performance on this task \cite{zhang2022motiondiffuse,dabral2022mofusion}. In these works, sequences of human poses are generated by a diffusion model conditioned on the output of a frozen CLIP text encoder. It is worth noting that these prior works model only a single human, while our framework supports multiple human agents and objects, and models their interactions with the environment.

\section{Method}
\label{sec:method}

\begin{figure*}
\begin{subfigure}{0.48\textwidth}
    \includegraphics[width=\textwidth]{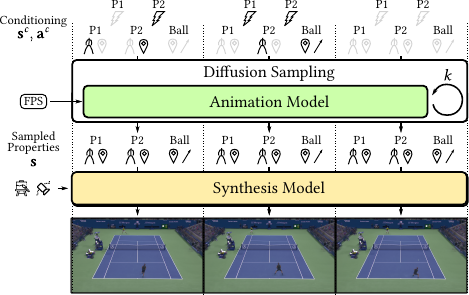}
    \caption{Model Overview}
    \label{fig:architecture_model_overview}
\end{subfigure}
\hfill
\begin{subfigure}{0.48\textwidth}
    \includegraphics[width=\textwidth]{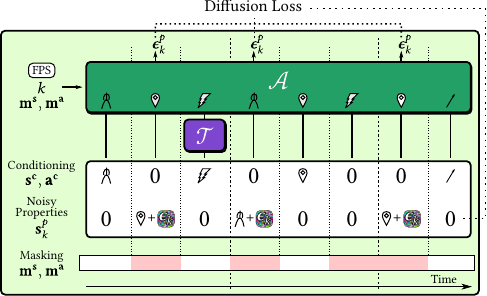}
    \caption{Animation model}
    \label{fig:architecture_animation_model}
\end{subfigure}
\hfill
\begin{subfigure}{0.99\textwidth}
    \includegraphics[width=\textwidth]{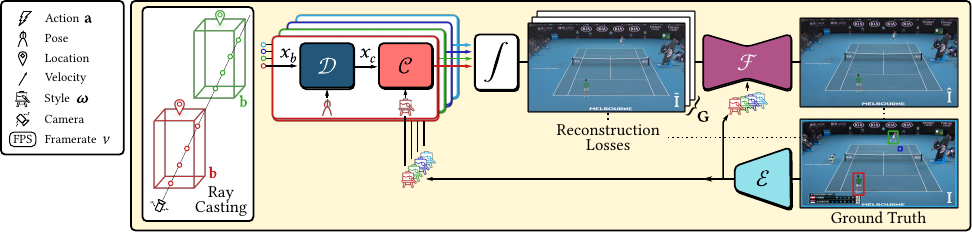}
    \caption{Synthesis model}
    \label{fig:architecture_synthesis_model}
\end{subfigure}

  \caption{(a) Overview of our framework. The \emph{animation model} produces states $\vecsequence$ based on user-provided conditioning signals, \change{or \emph{prompts},} $\vecsequencecond, \vectextactioncond$ that are rendered by the \emph{synthesis model}. (b) The diffusion-based animation model predicts noise $\vecdiffnoise_\difftimestep$ applied to the noisy states $\vecsequenceprednoise_\difftimestep$ conditioned on known states $\vecsequencecond$ and actions $\vectextactioncond$ with the respective masks $\vecmask, \vecmaskaction$, diffusion step $\difftimestep$ and framerate $\framerate$. The \emph{text encoder} $\nettext$ produces embedding for the textual actions, while the \emph{temporal model} $\netanimation$ performs noise prediction. (c) The synthesis model renders the current state using a composition of neural radiance fields, one for each object. A \emph{style encoder} $\netstyle$ extracts the appearance $\vecstyle$ of each object. Each object is represented in its canonical pose by $\netcanonical$ and deformations of articulated objects are modeled by the \emph{deformation model} $\netdeformation$. After integration and composition, the feature grid $\tensnerffeatures$ is rendered to the final image using the \emph{feature enhancer} $\netenhancer$.}
  \label{fig:architecture} 
\end{figure*} 

This section introduces our framework for the creation of \change{\emph{Promptable Game Models}} that allows the user to perform a range of dynamic scene editing tasks, \change{formulated as a set of conditioning \emph{prompts}}. 

We divide our \change{PGM} into two modules: a \emph{synthesis model} and an \emph{animation model}. The synthesis model generates an image given the representation of the environment state. The animation model, instead, aims at modeling the game's dynamics, with player actions and interactions, in the high-level space of the environment states.  Actions are modeled as text, which is an expressive, yet intuitive form of control for a wide range of tasks.
The overview of our framework is provided in Fig.~\ref{fig:architecture_model_overview}.

In more detail, our model defines the state of the entire environment as the combination of all individual object states. Consequently, each individual state is the set of the object properties such as the position of each object in the scene, their appearance, or their pose. Formally, the environment state at time $\seqtimestep$ can be represented by $\vecsequence_\seqtimestep \in \mathcal{\setstate}=(\mathbb{R}^{n_1} \times ... \times \mathbb{R}^{n_{\numproperties}})$, $\numproperties$ properties of variable length $n_i$ defined as the union of the properties of each object. This state representation captures all variable aspects of each object in the environment, thus it can be used by the synthesis model to generate the scene.

On the other hand, the animation model predicts the evolution of an environment in time, which is represented by the sequence of its states $\{\vecsequence_1,\vecsequence_2, 
\dots \vecsequence_{\numseqtimesteps}\} =\vecsequence \in \mathcal{\setstate}^{\numseqtimesteps}$, where $\numseqtimesteps$ is the length of the sequence. The model provides control over sequence generation with the help of user-defined conditioning signals, \change{or prompts,} that can take two forms: explicit
state manipulation and high-level text-based editing. With respect to the former, the user could change the position of the tennis ball at time step $\seqtimestep$, and the model would automatically adapt the position of the ball in other nearby states. As far as the latter is concerned, users could provide high-level text-based values of actions such as \emph{"The player takes several steps to the right and hits the ball with a backhand"} and the model would generate the corresponding sequence of states (see Fig.~\ref{fig:alternative_actions_qualitatives}). 
These generic actions in the form of text are central to enabling high-level, yet fine-grained control over the evolution of the environment. To train our framework we assume a dataset of camera-calibrated videos, where each video frame is annotated with the corresponding states $\vecsequence$ and actions $\vectextaction$. 

\subsection{Synthesis Model}
\label{sec:synthesis_model}
In this section, we describe the synthesis model that renders states from controllable viewpoints (see Fig.~\ref{fig:architecture_synthesis_model}). We build our model based on a compositional NeRF~\cite{Menapace2022PlayableEnvironments} framework which enables explicit control over the camera and represents a scene as a composition of different, independent objects. Thanks to the independent representation of objects, each object property is directly linked to an aspect of the respective object and can thus be easily controlled and manipulated. The compositional NeRF framework allows different, specialized NeRF architectures to be used for each object based on its type. To further improve quality, rather than directly rendering RGB images with the NeRF models, we render features and make use of a feature enhancer CNN to produce the RGB output. In order to represent objects with different appearances, we condition the NeRF and enhancer models on the style codes extracted with a dedicated style encoder \cite{Menapace2022PlayableEnvironments}. Our model is trained using reconstruction as the main guiding signal.

In Sec.~\ref{sec:nerf_overview}-\ref{sec:objs} we illustrate the main components of the synthesis module and in Sec.~\ref{sec:synthesis_training} we describe the training procedure.

\subsubsection{Scene Composition with NeRFs}
\label{sec:nerf_overview}
Neural radiance fields represent a scene as a radiance field, a 5D function parametrized as a neural network mapping the current position $\vecpoint$ and viewing direction $\vecdirection$ to density $\vecdensity$ and radiance $\veccolor$. 

To allow controllable generation of complex scenes, we adopt a compositional strategy where each object in the scene is modeled with a dedicated NeRF model \cite{Menapace2022PlayableEnvironments,mueller2022autorf,xu2022discoscene}.
The scene is rendered by sampling points independently for each object and querying the respective object radiance field $\netcanonical_i$. \change{The results for all objects are then merged and sorted by distance from the camera before being integrated.}

All objects are assumed to be described by a set of properties whose structure depends on the type of object, \emph{e.g.} a player, the ball, the background. We consider the following properties:
\begin{itemize}
\item \emph{Object location}.  Each object is contained within an axis-aligned bounding box $\vecbboxthree_i$ which is defined by size and position. In the case of the ball, we additionally consider its velocity to model blur effects (Sec.~\ref{sec:objs}). 
\item \emph{Object style}. All objects have an appearance that may vary in different sequences, thus we introduce a style code $\vecstyle_i$ as an additional property for all objects. Since it is difficult to define such style information a priori, we assume it to be a latent variable and learn it jointly during training.
\item \emph{Object pose}. Articulated objects such as humans require additional properties to model varying poses. We model the deformation of articulated objects as a kinematic tree with ${\numjoints}_i$ joints and consider as object properties the rotation $\tensrotation$ and translation $\vectranslation$ parameters associated with each joint (Sec.~\ref{sec:deformation_modeling}). 
\end{itemize}
From now on, we drop the object index $i$ to simplify notation.

\subsubsection{Style Encoder}
\label{sec:style_encoder}
Representing the appearance of each object is challenging since it changes based on the type of object and illumination conditions. We treat the style $\vecstyle$ for each object as a latent variable that we regress using a convolutional style encoder $\netstyle$. Given the current video frame $\tensimage$ with $\numobjects$ objects, we compute 2D bounding boxes ${\vecbboxtwo}$ for each object. First, a set of residual blocks is used to extract frame features which are later cropped around each object according to ${\vecbboxtwo}$ using RoI pooling \cite{girshick2013roipool}. Later, a series of convolutional layers with a final projection is used to predict the style code $\vecstyle$ from the cropped feature maps.

\subsubsection{Volume Modeling for Efficient Sampling}
\label{sec:canonical_volume}

Radiance fields are commonly parametrized using MLPs~\cite{mildenhall2020nerf} but such representation requires a separate MLP evaluation for each sampled point, making it computationally challenging to train high-resolution models.
To overcome such issue, we model the radiance field $\netcanonical$ of each object in a canonical space using two alternative parametrizations.

For three-dimensional objects, 
we make use of a voxel grid parametrization \cite{fridovich2022plenoxels,weng2022humannerf}. Starting from a fixed noise tensor $\tensvoxelsmall \in \mathbb{R}^{\numfieldfeatsmall \times \numvoxelheightsmall \times \numvoxelwidthsmall \times \numvoxeldepthsmall}$, a series of 3D convolutions produces a voxel $\tensvoxel \in \mathbb{R}^{\numfieldfeat + 1 \times \numvoxelheight \times \numvoxelwidth \times \numvoxeldepth}$ containing the features and density associated to each point in the bounded space. Here, $\numfieldfeatsmall$ and $\numfieldfeat$ represent the number of features, while $\numvoxelheight$, $\numvoxelwidth$ and $\numvoxeldepth$ represent the size of the voxel. Given a point in the object canonical space $\vecpointcanon$, the associated features and density $\vecdensity$ are retrieved using trilinear sampling on $\tensvoxel$. To model the different appearance of each object, we adopt a small MLP conditioned on the style $\vecstyle$ to produce a stylized feature with the help of weight demodulation~\cite{karras2019stylegan2}. 

For two-dimensional objects such as planar scene elements, we make use of a similar parametrization  where a fixed 2D noise tensor $\tensplanesmall \in \mathbb{R}^{\numfieldfeatsmall \times \numplaneheightsmall \times \numplanewidthsmall}$ is mapped to a plane of features $\tensplane \in \mathbb{R}^{\numfieldfeat \times \numplaneheight \times \numplanewidth}$ using a series of 2D convolutions.  
Given a ray $\ray$, we compute the intersection point $\vecpoint$ between the plane and the ray which is used to sample $\tensplane$ using bilinear sampling. Similarly to the voxel case, a small MLP is used to model object appearance according to $\vecstyle$. We assume planes to be fully opaque and assign a fixed density value $\vecdensity$ to each sample. Thanks to this representation, a single point per ray is sufficient to render the object.

\subsubsection{Deformation Modeling}
\label{sec:deformation_modeling}
Since the radiance field $\netcanonical$ alone supports only rendering of rigid objects expressed in a canonical space, to render articulated objects such as humans we introduce a deformation model $\netdeformation$. Given an articulated object, we assume its kinematic tree is known and that the transformation $[\tensrotation_j|\vectranslation_j]$ from each joint $j \in 1,...,\numjoints$ to the parent joint is part of the object's properties.  We then implement a deformation procedure based on linear blend skinning (LBS) \cite{lewis2000pose} and inspired by HumanNeRF \cite{weng2022humannerf} that employs the joint transformations and a learned volume of blending weights $\tensblendweights \in \mathbb{R}^{\numjoints + 1 \times \numblendweightsheight \times \numblendweightswidth \times \numblendweightsdepth}$ to associate each point in the bounding box of the articulated object to the corresponding one in the canonical volume. We present additional details in \apref{ap:deformation_modeling}.

\subsubsection{Enhancer}
\label{sec:enhancer}
NeRF models are often parametrized to output radiance $\veccolor \in \mathbb{R}^3$ and directly produce an image. However, we find that such approach struggles to produce correct shading of the objects, with details such as shadows being difficult to synthesize. Also, to improve the computational efficiency of the method, we sample a limited number of points per ray that may introduce subtle artifacts in the geometry. To address these issues, we parametrize the model $\netcanonical$ to output features where the first three channels represent radiance and the subsequent represent learnable features. Then, we produce a feature grid $\tensnerffeatures \in \mathbb{R}^{\numfieldfeat \times \numimgheight \times \numimgwidth}$ and an RGB image $\tensimagergb \in \mathbb{R}^{3 \times \numimgheight \times \numimgwidth}$. We introduce an enhancer network $\netenhancer$ modeled as a UNet~\cite{ronneberger2015unet} architecture interleaved with weight demodulation layers~\cite{karras2019stylegan2} that maps $\tensnerffeatures$ and the style codes $\vecstyle$ to the final RGB output $\tensimagerec \in \mathbb{R}^{3 \times \numimgheight \times \numimgwidth}$.

\subsubsection{Object-specific rendering}
\label{sec:objs}
Our compositional approach allows the use of object-specific techniques. In particular, in the case of tennis, we detail in \apref{ap:object_specific_techniques} how we can apply dedicated procedures to enhance the rendering quality of the ball, the racket, and the 2D user interfaces such as the scoreboards. The rendering of the tennis ball is treated specially to render the blur that occurs in real videos in the case of fast-moving objects. The racket can be inserted in a post-processing stage to compensate for the difficulty of NeRFs to render thin, fast-moving objects. Finally, the UI elements are removed from the scene since they do not behave in a 3D consistent manner. For Minecraft, we describe how the scene skybox is modeled.

\subsubsection{Training}
\label{sec:synthesis_training}
We train our model using reconstruction as the main driving signal. Given a frame $\tensimage$ and reconstructed frame $\tensimagerec$, we use a combination of L2 reconstruction loss and the perceptual loss of Johnson \etal \cite{johnson2016perceptual} as our training loss. To minimize the alterations introduced by the enhancer and improve view consistency, we impose the same losses between $\tensimage$ and $\tensimagergb$, the output of the model before the feature enhancer. All losses are summed without weighting to produce the final loss term. To minimize GPU memory consumption, instead of rendering full images, we impose the losses on sampled image patches instead~\cite{Menapace2022PlayableEnvironments}.

We train all the components of the synthesis model jointly using Adam \cite{kingma2014adam} for 300k steps with batch size 32. We set the learning rate to $1e-4$ and exponentially decrease it to $1e-5$ at the end of training.
 The framework is trained on videos with 1024x576px resolution. We present additional details in \apref{ap:implementation_details_synthesis} and in \apref{ap:training_details_synthesis}, \change{and discuss inference details in \apref{ap:inference_details}}.

\subsection{Animation Model}
\label{sec:animation_model}

In this section, we describe the animation model (see Fig.~\ref{fig:architecture_animation_model}), whose task is that of generating sequences of states $\vecsequence \in \mathcal{\setstate}^{\numseqtimesteps}$ according to user inputs. The animation model allows users to specify conditioning signals, \change{or prompts,} in two forms. First, conditional signals can take the form of values that the user wants to impose on some object properties in the sequence, such as the player position at a certain time step.
This signal is represented by a sequence $\vecsequencecond \in \mathcal{\setstate}^{\numseqtimesteps}$. This form of conditioning allows fine control over the sequence to generate but requires directly specifying values of properties. Second, to allow high-level, yet granular control over the sequence, we introduce actions in the form of text $\vectextactioncond \in {\settext}^{\numactions \times \numseqtimesteps}$ that specify the behavior of each of the $\numactions$ actionable objects at each timestep in the sequence, where $\settext$ is the set of all strings of text.
To maximize the flexibility of the framework, we consider all values in $\vecsequencecond$ and $\vectextactioncond$ to be optional, thus we introduce their respective masks $\vecmask \in \{0,1\}^{\numproperties \times \numseqtimesteps}$ and $\vecmaskaction \in \{0,1\}^{\numactions \times \numseqtimesteps}$ that are set to 1 when the respective conditioning signal is present. We assume elements where the mask is not set to be equal to 0. 
The animation model predicts $\vecsequencepred \in \mathcal{\setstate}^{\numseqtimesteps}$ conditioned on $\vecsequencecond$ and $\vectextactioncond$ such that:
\begin{equation}
\label{eq:sequence_composition}
\vecsequence = \vecsequencepred + \vecsequencecond,
\end{equation}
where we consider the entries in $\vecsequencepred$ and $\vecsequencecond$ to be mutually exclusive, \emph{i.e.} an element of $\vecsequencepred$ is 0 if the corresponding conditioning signal in $\vecsequencecond$ is present according to $\vecmask$. Note that the prediction of actions is not necessary, since $\vecsequence$ is sufficient for rendering.
 
 We adopt a temporal model $\netanimation$ based on a non-autoregressive masked transformer design and leverage the knowledge of a pretrained language model in a text encoder $\nettext$ to model action conditioning information \cite{han2022show}. The masked design provides support for the optional conditioning signals and is trained using masked sequence modeling, where we sample $\vecmask$ and $\vecmaskaction$ according to various strategies that emulate desired inference tasks.

In Sec.~\ref{sec:text_encoder} we define our text encoder, Sec.~\ref{sec:temporal_model} defines the diffusion backbone, and in Sec.~\ref{sec:animation_training} we describe the training procedure.

\subsubsection{Text Encoder}
\label{sec:text_encoder}

We introduce a text encoder $\nettext$ that encodes textual actions into a sequence of fixed-size text embeddings:
\begin{equation}
\label{eq:text_encoder}
\vecactionemb = \nettext(\vectextactioncond) \in \mathbb{R}^{\numactions \times \numseqtimesteps \times \numactionembsize},
\end{equation}
where $\numactionembsize$ is the size of the embedding for the individual sentence. Given a textual action, we leverage a pretrained T5 text model \cite{raffel2022exploring} $\nettextenc$ that tokenizes the sequence and produces an output feature for each token. Successively, a feature aggregator $\nettextagg$ modeled as a transformer encoder \cite{vaswani2017attention} produces the aggregated text embedding from the text model features. To retain existing knowledge into $\nettextenc$, we keep it frozen and only train the feature aggregator $\nettextagg$.

\subsubsection{Temporal Modeling}
\label{sec:temporal_model}

In this section, we introduce the temporal model $\netanimation$ that predicts the sequence of states $\vecsequence$ conditioned on known state values $\vecsequencecond$, action embeddings $\vecactionemb$, and the respective masks $\vecmask$ and $\vecmaskaction$. Since only unknown state values need to be predicted, the model predicts $\vecsequencepred$ and the complete sequence of states is obtained as $\vecsequence = \vecsequencepred + \vecsequencecond$, following Eq.~\eqref{eq:sequence_composition}. Diffusion models have recently shown state-of-the-art performance on several tasks closely related to our setting such as sequence modeling \cite{tashiro2021csdi} and text-conditioned human motion generation \cite{dabral2022mofusion,zhang2022motiondiffuse}. Thus, we follow the DDPM \cite{ho2020ddpm} diffusion framework, and we frame the prediction of $\vecsequencepred=\vecsequencepred_0$ as a progressive denoising process $\vecsequencepred_0,...,\vecsequencepred_\numdifftimesteps$, where we introduce the diffusion timestep index $\difftimestep \in {0,...,\numdifftimesteps}$. The temporal model $\netanimation$ acts as a noise estimator that predicts the Gaussian noise $\vecdiffnoise_\difftimestep$ in the noisy sequence of unknown states $\vecsequenceprednoise_\difftimestep$ at diffusion timestep $\difftimestep$:
\begin{equation}
\label{eq:temporal_model}
\vecdiffnoisepred_\difftimestep = \netanimation(\vecsequenceprednoise_\difftimestep|\vecsequencecond,\vecactionemb,\vecmask,\vecmaskaction,\difftimestep).
\end{equation}
An illustration of the proposed diffusion model is shown in Fig.~\ref{fig:architecture_animation_model}. 

We realize $\netanimation$ using a transformer encoder \cite{vaswani2017attention}. To prepare the transformer's input sequence, we employ linear projection layers $\netprojection$ with separate parameters for each object property. Since corresponding entries in $\vecsequenceprednoise_\difftimestep$ and $\vecsequencecond$ are mutually exclusive, we only consider the one that is present as input to the transformer and we employ different projection parameters to enable the model to easily distinguish between the two.
An analogous projection is performed for $\vecactionemb$ and, subsequently, the projection outputs for states and actions are concatenated into a single sequence $\vecsequenceemb \in \mathbb{R}^{\numproperties+\numactions \times \numseqtimesteps \times \numembsize}$, which constitutes the input to the transformer.
An output projection layer with separate weights for each object property produces the prediction $\vecdiffnoisepred_\difftimestep$ at the original dimensionality. 
To condition the model on the diffusion time-step $\difftimestep$, we introduce a weight demodulation layer \cite{karras2019stylegan2} after each self-attention and feedforward block \cite{zhang2022motiondiffuse}. 

To model long sequences while keeping reasonable computational complexity and preserving the ability to model long-term relationships between sequence elements, it is desirable to build the sequences using states sampled at a low framerate. However, this strategy would not allow the model to generate content at the original framerate and would prevent it from understanding dynamics such as limb movements that are clear only when observing sequences sampled at high framerates. To address this issue, we use the weight demodulation layers to further condition our model on the sampling framerate $\framerate$ to enable a progressive increase of the framerate at inference time (see \apref{ap:high_framerate_generation}).

\subsubsection{Training}
\label{sec:animation_training}
To train our model, we sample a sequence $\vecsequence$ with corresponding actions $\vectextaction$ from a video in the dataset at a uniformly sampled framerate $\framerate$. Successively, we obtain masks $\vecmask$ and $\vecmaskaction$ according to masking strategies we detail in \apref{ap:training_details_animation}. The sequence for training are obtained following $\vecsequencepred_0 = \vecsequence \odot (1-\vecmask)$ and $\vecsequencecond = \vecsequence \odot \vecmask$, and actions as $\vectextactioncond = \vectextaction \odot \vecmaskaction$, where 
$\odot$ denotes the Hadamard product.

We train our model by minimizing the DDPM \cite{ho2020ddpm} training objective:
\begin{equation}
\label{eq:diffusion_loss}
\mathbb{E}_{\difftimestep \sim \mathcal{U}(1,\numdifftimesteps),\epsilon \sim \mathcal{N}(0,I)}||\vecdiffnoisepred_\difftimestep - \vecdiffnoise_\difftimestep||,
\end{equation}
where $\vecdiffnoisepred_\difftimestep$ is the noise estimated by the temporal model $\netanimation$ according to Eq.~\eqref{eq:temporal_model}.
Note that the loss is not applied to positions in the sequence corresponding to conditioning signals \cite{tashiro2021csdi}.

Our model is trained using the Adam \cite{kingma2014adam} optimizer with a learning rate of $1e-4$, cosine schedule, and with 10k warmup steps. We train the model for a total of 2.5M steps and a batch size of 32. We set the length of the training sequences to $\numseqtimesteps=16$. The number of diffusion timesteps is set to $\numdifftimesteps=1000$ and we adopt a linear noise schedule \cite{ho2020ddpm}. Additional details are presented in \apref{ap:implementation_details_animation} \change{and \apref{ap:inference_details}}.

\section{Applications}
\label{sec:applications}

\begin{figure*}
\includegraphics[width=0.96\textwidth]{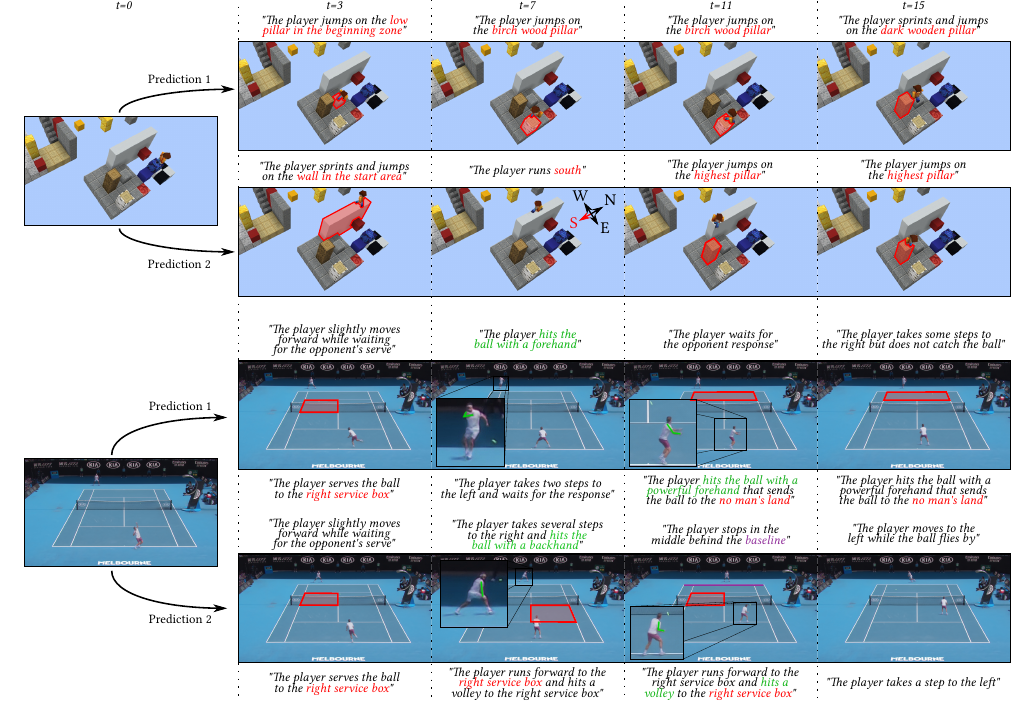}
  \caption{Different sequences predicted on the Tennis and Minecraft datasets starting from the same initial state and altering the text conditioning. Our model moves players and designates shot targets using domain-specific referential language (eg. \emph{"right service box"}, \emph{"no man's land"}, \emph{"baseline"}). The model supports fine-grained control over the various tennis shots using technical terms (eg. \emph{``forehand''}, \emph{``backhand''}, \emph{``volley'')}.}
  \label{fig:alternative_actions_qualitatives}
\end{figure*}

\begin{figure*}
\includegraphics[width=0.95\textwidth]{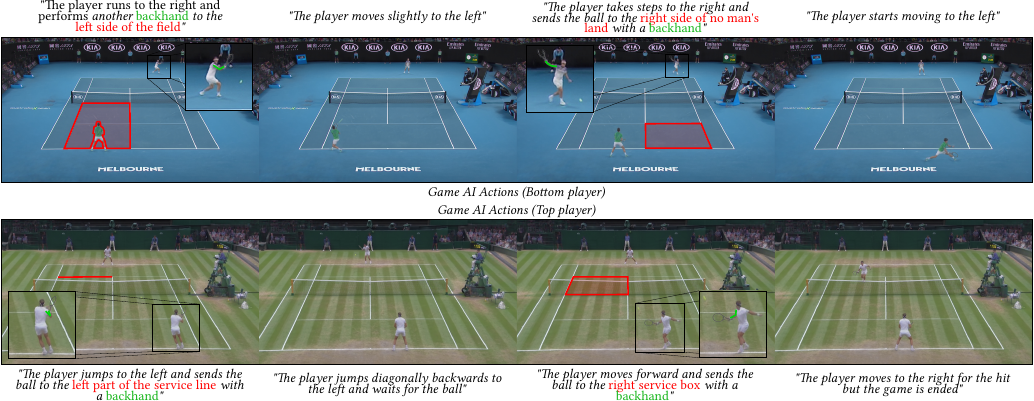}
  \caption{Sequences generated by specifying actions for one of the players and letting the model act as the game AI and take control of the opponent. The game AI successfully responds to the actions of the player by running to the right (see top sequence) or towards the net (see bottom sequence), following two challenging shots of the user-controlled player.}
  \label{fig:opponent_control}
\end{figure*}

\begin{figure*}
\includegraphics[width=\textwidth]{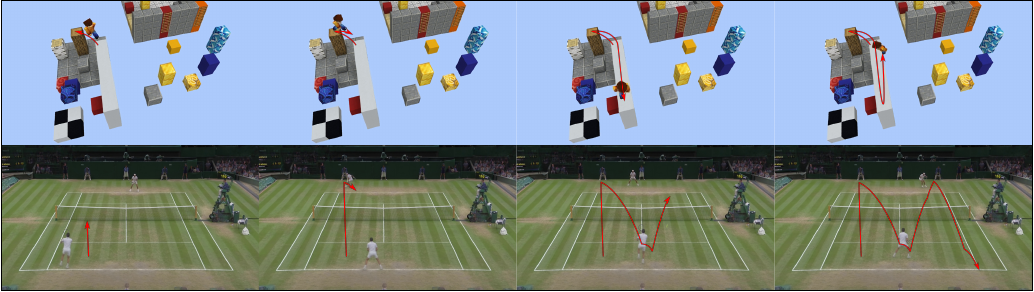}
  \caption{Sequences generated without any user conditioning signal. The actions of all players are controlled by the model that acts as the game AI. In tennis, the players produce a realistic exchange, with the bottom player advancing aggressively toward the net and the top player defeating him with a shot along the right sideline. The Minecraft player and tennis ball trajectories are highlighted for better visualization.}
  \label{fig:random_player_movement}
\end{figure*}

\begin{figure*}
\includegraphics[width=\textwidth]{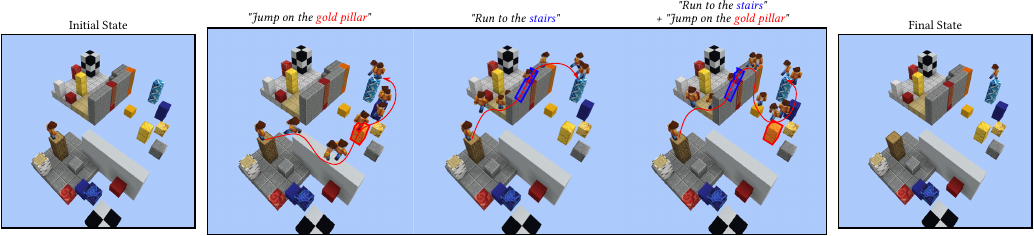}
  \caption{Given an initial and a final state, we generate all the states in between. We repeat the generation multiple times conditioning it using different actions indicating the desired intermediate waypoints.}
  \label{fig:minecraft_navigation}
\end{figure*}

\begin{figure*}
\includegraphics[width=\textwidth]{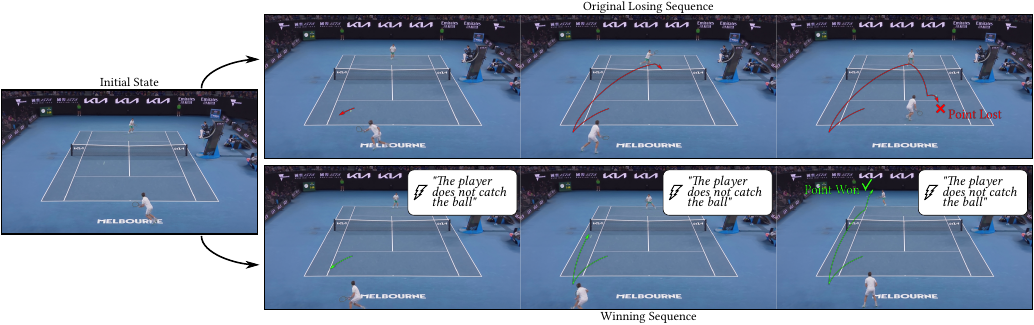}
  \caption{Given a sequence where the bottom player loses (see top), we ask the model to modify it such that the bottom player wins instead (see bottom). To do so, we condition the top player on the action \emph{"The player does not catch the ball"}. While in the original sequence the bottom player aims its response to the center of the field where the opponent is waiting, the model now successfully generates a winning set of moves for the bottom player that sends the ball along the left sideline, too far for the top player to be reached.}
  \label{fig:how_to_win}
\end{figure*}

Our framework enables a series of applications that are unlocked by its expressive state representation, the possibility to render it using a 3D-aware synthesis model, and the ability to generate sequences of states with an animation model that understands the game dynamics and can be conditioned on a wide range of signals 
In the following, we demonstrate a set of selected applications.

Our state representation is modular, where the style is one of the components. Style swapping is enabled by swapping the style of the desired object $\vecstyle$ in the original image with the one from a target image. Similarly to a \change{traditional} game engine, our synthesis model renders the current state of the environment from a user-defined perspective. This enables our model to perform novel view synthesis. We show in \apref{ap:applications} examples of both these capabilities.

We now show a set of applications enabled by the animation model. In Fig.~\ref{fig:alternative_actions_qualitatives}, we show results for generating different sequences using textual actions starting from a common initial state. Thanks to the textual action representation, it is possible to gain fine control over the generated results and to make use of referential language. 

Our animation model, however, is not limited 
to generate sequences given step-by-step actions. Thanks to its understanding of the game's dynamics, the model can tackle more complex tasks such as modeling an opponent against which a user-controlled player can play (see Fig.~\ref{fig:opponent_control}), or even controlling all players without user intervention (see Fig.~\ref{fig:random_player_movement}), in a way similar to a ``game AI''.

The animation model also unlocks the "director's mode", where the user can generate sequences by specifying \change{prompts} consisting in a desired set of high-level constraints or goals. The model is able to reason on actions to find a solution satisfying the given constraints. As a first example, Fig.~\ref{fig:minecraft_navigation} demonstrates results for a navigation problem, where the user specifies a desired initial and final player position in the scene, and the model devises a path between them. 
Notably, the user can also constrain the solution on intermediate waypoints by means of natural language. 
As a second example, Fig.~\ref{fig:how_to_win} shows that the model is capable of devising strategies to defeat an opponent. Given an original sequence where the player commits a mistake and loses, the model can devise which actions the player should have taken to win. 
Notably, these model capabilities are learned by just observing sequences annotated with textual actions.

\section{Evaluation}
\label{sec:evaluation}
In this section, we introduce our Tennis and Minecraft datasets (Sec.~\ref{sec:datasets}), describe our experimental protocol (Sec.~\ref{sec:evaluation_protocol}), and perform evaluation of both the synthesis model (Sec.~\ref{sec:synthesis_evaluation}) and the animation model (Sec.~\ref{sec:animation_evaluation}). Additional evaluation results are shown in \apref{ap:evaluation}.

\begin{figure*}
\includegraphics[width=\textwidth]{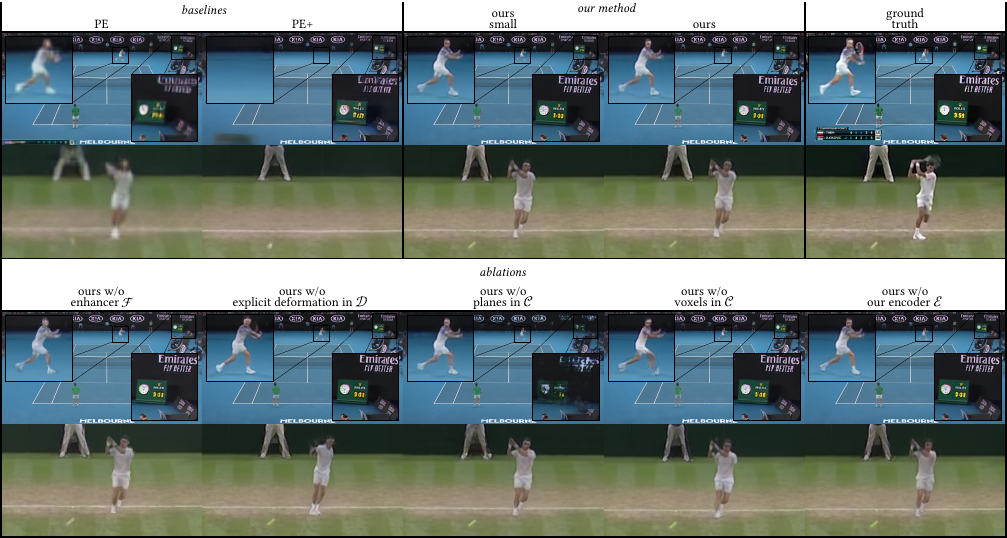}
  \caption{Synthesis model qualitative results on the Tennis dataset. Compared to PE \cite{Menapace2022PlayableEnvironments}, our model generates sharper players and static scene elements. Our ablation study shows corruption of the player geometry when voxels or our deformation model are not used. When removing our canonical plane representation, static scene elements appear blurry. When our feature enhancer is removed, the model does not generate shadows and players lose quality.}
  \label{fig:synthesis_ablation_qualitatives}
\end{figure*}

\subsection{Datasets}
\label{sec:datasets}
We collect two datasets to evaluate our method. Both datasets and the employed data collection tools are publicly available. In the following, we describe their structure and the available annotations.

\subsubsection{Tennis dataset}
We collect a dataset of broadcast tennis matches starting from the videos in \cite{Menapace2022PlayableEnvironments}. The dataset depicts matches between two professional players from major tennis tournaments, captured with a single, static bird's eye camera.

To enable the construction of \change{PGMs}, we collect a wide range of annotations with a combination of manual and automatic methods (see \apref{ap:tennis_dataset_collection}):
\begin{itemize}
\item For each frame, we perform camera calibration.
\item For each of the two players, we perform tracking and collect full SMPL \cite{loper2015smpl} body parameters. Note that in our work we only use a subset of the parameters: rotation and translation associated with each joint, and the location of the root joint in the scene. 
\item For each player and frame, we manually annotate textual descriptions of the action being performed. We structure captions so that each includes information on where and how the player is moving, the particular type of tennis shot being performed, and the location where the shot is aimed (see \apref{ap:dataset_samples}). Captions make use of technical terms to describe shot types and field locations. In contrast to other video-text datasets that contain a single video-level \cite{bain2021webvid10m} or high-level action descriptions weakly aligned with video content \cite{miech19howto100m}, the captions in our dataset are separate for each object and constitute a fine-grained description of the actions taking place in the frame.
\item For the ball, we perform 3D tracking and provide its position in the scene and its velocity vector indicating the speed and direction of movement. 
\end{itemize}

We collect 7112 video sequences in 1920x1080px resolution and 25fps starting from the videos in \cite{Menapace2022PlayableEnvironments} for a total duration of 15.5h. The dataset features 1.12M fully-annotated frames and 25.5k unique captions with 915 unique words. We highlight key statistics of the dataset and show samples in \apref{ap:datasets}.

We note that broadcast Tennis videos are monocular and do not feature camera movements other than rotation, thus the dataset does not make it possible to recover the 3D geometry of static objects \cite{Menapace2022PlayableEnvironments}. 

\subsubsection{Minecraft dataset}
We collect a synthetic dataset from the Minecraft video game. This dataset depicts a player performing a series of complex movements in a static Minecraft world that include walking, sprinting, jumping, and climbing on various world structures such as platforms, pillars, stairs, and ladders. A single, monocular camera that slowly orbits around the scene center is used to capture the scenes. We collect a range of synthetic annotations using a game add-on we develop starting from \cite{replaymod}:
\begin{itemize}
\item Camera calibration for each frame.
\item Player rotation and translation parameters associated with each joint in the Minecraft kinematic tree format, and the location of the root joint in the scene (see \apref{ap:minecraft_skeleton_format}).
\item A synthetically-generated text caption describing the action being performed by the player. We assign varied, descriptive names to each element of the scene and build captions that describe scene elements or directions towards which the player is moving. Additionally, our captions capture how movement is happening i.e. by jumping, sprinting, walking, climbing, or falling. We adopt a stochastic caption generation procedure that generates multiple alternative captions for each frame.
\end{itemize}

A total of 61 videos are collected in 1024x576px resolution and 20fps for a total duration of 1.21h. The dataset contains 68.5k fully annotated frames and 1.24k unique captions with 117 unique words. We highlight key statistics for the dataset in \apref{ap:datasets}.

\subsection{Evaluation Protocol}
\label{sec:evaluation_protocol}
We evaluate the synthesis and the animation models separately, following a similar evaluation protocol. We divide the test dataset into non-overlapping sequences of 16 frames sampled at 5fps and 4fps respectively for the Minecraft and Tennis datasets and make use of the synthesis or animation model to reconstruct them. In the case of the synthesis model, we directly reconstruct the video frames and compute the following metrics:
\begin{itemize}
\item \textit{LPIPS} \cite{zhang2018unreasonable} is a standard metric for evaluating the reconstruction quality of the generated images
\item \textit{FID} \cite{heusel2017advances} is a widely-used metric for image generation quality
\item \textit{FVD} \cite{unterthiner2018towards} is a standard metric for assessing the quality of generated videos
\item \textit{Average Detection Distance (ADD)} \cite{menapace2021pvg} measures the average distance in pixels between the bounding box centers of ground truth bounding boxes and bounding boxes obtained from the generated sequences through a pretrained detector
\cite{ren2015faster}
\item \textit{Missing Detection Rate (MDR)} \cite{menapace2021pvg}  estimates the rate of bounding boxes that are present in the ground truth, but that are missing in the generated videos
\end{itemize}

For the animation model, we evaluate reconstruction of the object properties. Note that different strategies for masking affect the behavior of the model and the nature of the reconstruction task, thus we separately evaluate different masking configurations corresponding to different inference tasks. We compute metrics that address both the fidelity of the reconstruction and the realism of the produced sequences:
\begin{itemize}
\item \textit{L2} computes the fidelity of the reconstruction by measuring the distance between the ground truth and reconstructed object properties along the sequence
\item \textit{Fréchet Distance (FD)} \cite{frechet1957distance} measures the realism of each object property by computing the Fréchet Distance between the distribution of real sequences of a certain object property and of generated ones.
\end{itemize}

We select different reconstruction tasks for evaluation:
\begin{itemize}
\item \textit{Video prediction conditioned on actions} consists in reconstructing the complete sequence starting from the initial state while the actions are specified for all timesteps. This setting corresponds to the evaluation setting of \cite{Menapace2022PlayableEnvironments}.
\item \textit{Unconditioned video prediction} consists in reconstructing the complete sequence starting from the first state only.
\item \textit{Opponent modeling} consists in reconstructing the object properties of an unknown player, based on the state of the other player, with actions specified only on the known player. Good performance in this task indicates the ability to model an opponent against which a user can play.
\item \textit{Sequence completion} consists in reconstructing a sequence where 8 consecutive states are missing. No actions are specified for the missing states. Good performance in this task indicates ability in reasoning on how it is possible to reach a certain goal state starting from the current one.
\end{itemize}

\subsection{Synthesis Model Evaluation}
\label{sec:synthesis_evaluation}
In this section, we evaluate the performance of the synthesis model.

\subsubsection{Comparison to Baselines}
We evaluate our method against Playable Environments (PE) \cite{Menapace2022PlayableEnvironments}, the work most related to ours in that it builds a controllable 3D environment representation that is rendered with a compositional NeRF model where the position of each object is given and pose parameters are treated as a latent variable. Since the original method supports only outputs at 512x288px resolution, we produce baselines trained at both 512x288px and 1024x576px resolution which we name PE and PE+ respectively. For a fair comparison, we also introduce in the baselines our same mechanism for representing ball blur and train a variant of our model using the same amount of computational resources as the baselines (Ours Small).

Results of the comparison are shown in Tab.~\ref{table:synthesis_merged}, while qualitative results are shown in Fig.~\ref{fig:synthesis_ablation_qualitatives}. 
Our method scores best in terms of LPIPS, ADD and MDR. Compared to PE+, our method produces significantly better FID and FVD scores. As shown in Fig.~\ref{fig:synthesis_ablation_qualitatives}, PE and PE+ produce checkerboard artifacts that are particularly noticeable on static scene elements such as judge stands, while our method produces sharp details. We attribute this difference to our ray sampling scheme and feature enhancer design that, in contrast to PE, do not sample rays at low resolution and perform upsampling, but rather directly operate on high resolution. In addition, thanks to our deformation and canonical space modeling strategies, and higher resolution, our method produces more detailed players with respect to PE, where they frequently appear with missing limbs and blurred clothing. Finally, our model produces a realistic ball, while PE struggles to correctly model small objects, presumably due to its upsampling strategy that causes rays to be sampled more sparsely and thus do not intersect with the ball frequently enough to correctly render its blur effect.

\begin{table}[t]
\caption{Comparison with baselines and ablation of the synthesis model. MDR in \%, ADD in pixels. Note that FID and FVD are computed on images downscaled to the feature extractor training resolution, thus blurriness in the PE baseline caused by its reduced resolution is not captured by these metrics. LPIPS correctly reflects lack of sharpness in the PE results (see Fig.~\ref{fig:synthesis_ablation_qualitatives}). $\dag$ denotes output in 512x288px rather than 1024x576px resolution.}

\label{table:synthesis_merged}

\begin{center}

\footnotesize
\begin{tabular}{lccccc}
\toprule
 \multicolumn{1}{c}{\emph{Tennis}} & LPIPS$\downarrow$  & FID$\downarrow$ & FVD$\downarrow$ & ADD$\downarrow$ & MDR$\downarrow$\\
\midrule

PE$^\dag$ \cite{Menapace2022PlayableEnvironments} & 0.188 & \cellfirst11.5 & \cellfirst349 & 3.74 & 0.200  \\
PE+ \cite{Menapace2022PlayableEnvironments} & 0.232 & 40.4 & 2432 & 132.3 & 49.7  \\
\midrule
w/o enhancer $\netenhancer$ & \cellthird0.167 & 15.6 & 570 & \cellthird3.02 & 0.0728  \\
w/o explicit deformation in $\netdeformation$ & \cellsecond0.156 & 13.3 & 524 & 3.10 & 0.0587  \\
w/o planes in $\netcanonical$ & 0.241 & 30.4 & 1064 & \cellsecond2.94 & 0.0611  \\
w/o voxels in $\netcanonical$ & 0.170 & 17.1 & 757 & 3.03 & \cellfirst0.0399  \\
w/o our encoder $\netstyle$ & 0.174 & 15.0 & 600 & 3.18 & 0.0564  \\
\midrule
Ours Small & \cellsecond0.156 & \cellthird13.4 & \cellthird523 & \cellfirst2.88 & \cellthird0.0470  \\
Ours & \cellfirst0.152 & \cellsecond12.8 & \cellsecond516 & \cellfirst2.88 & \cellsecond0.0423  \\

\toprule
 \multicolumn{1}{c}{\emph{Minecraft}} & LPIPS$\downarrow$  & FID$\downarrow$ & FVD$\downarrow$ & ADD$\downarrow$ & MDR$\downarrow$\\
\midrule

PE$^\dag$ \cite{Menapace2022PlayableEnvironments} & \cellthird0.0235 & \cellthird13.9 & \cellthird21.5 & \cellthird5.77 & \cellthird0.0412  \\
PE+ \cite{Menapace2022PlayableEnvironments} & 0.0238 & 15.5 & 51.7 & 120.6 & 0.939  \\
\midrule
Ours Small & \cellsecond0.00996 & \cellsecond3.56 & \cellsecond8.83 & \cellsecond2.02 & \cellsecond0.0529  \\
Ours & \cellfirst0.00814 & \cellfirst2.81 & \cellfirst7.08 & \cellfirst1.98 & \cellfirst0.0508  \\

\bottomrule

\end{tabular}
\end{center}

\end{table}

\subsubsection{Ablation}
To validate our design choices, we produce several variations of our method, each produced by removing one of our proposed architectural elements: we remove the enhancer $\netenhancer$ and directly consider $\tensimagergb$ as our output; we remove the explicit deformation modeling procedure in $\netdeformation$ of Sec.~\ref{sec:deformation_modeling} and substitute it with an MLP directly predicting the deformation using a learnable pose code as in \cite{Menapace2022PlayableEnvironments,tretschk2021nonrigid}; we remove the plane-based canonical volume representation in $\netcanonical$ for planar objects and use an MLP instead; we remove the voxel-based volume representation in $\netcanonical$ and use an MLP instead; we substitute our style encoder $\netstyle$ with an ad-hoc encoder for each object in the scene, following \cite{Menapace2022PlayableEnvironments}.

We perform the ablation on the Tennis dataset and show results in Tab.~\ref{table:synthesis_merged} and Fig.~\ref{fig:synthesis_ablation_qualitatives}. To reduce computation, we train the ablation models using the same hyperparameters as the ``Ours Small'' model.

When removing the enhancer $\netenhancer$, our model produces players with fewer details and does not generate shadow effects below players (see first row in Fig.~\ref{fig:synthesis_ablation_qualitatives}). When our deformation modeling procedure is not employed, the method produces comparable LPIPS, FID, and FVD scores, but an analysis of the qualitatives shows that players may appear with corrupted limbs (see last row in Fig.~\ref{fig:synthesis_ablation_qualitatives}). In addition, the use of such learned pose representation would reduce the controllability of the synthesis model with respect to the use of an explicit kinematic tree. When plane-based or voxel-based canonical modeling is removed, we notice artifacts in the static scene elements, such as corrupted logos, and in the players, such as detached or doubled limbs. Finally, when we replace our style encoder design with the one of \cite{Menapace2022PlayableEnvironments}, we notice fewer details in scene elements.

\subsection{Animation Model Evaluation}
\label{sec:animation_evaluation}
In this section, we evaluate the performance of the animation model.

\begin{figure}
\includegraphics{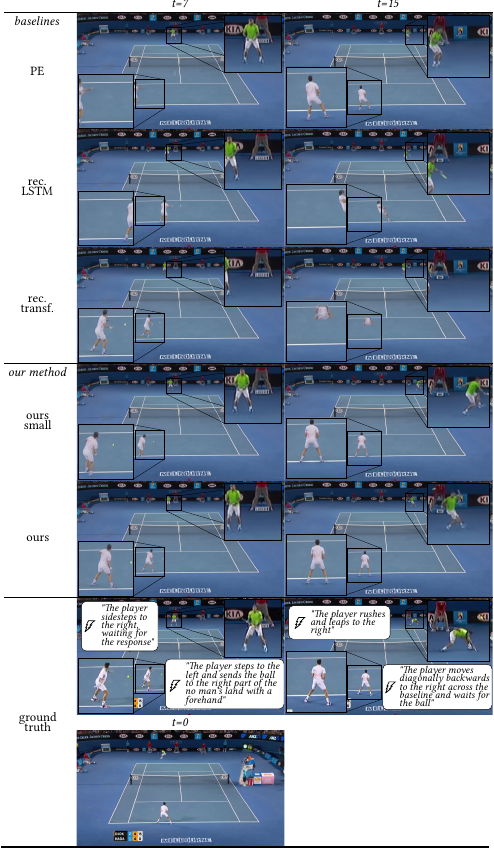}
  \caption{Qualitative results on the Tennis dataset. Sequences are produced in a video prediction setting that uses the first frame object properties and all actions as conditioning. The location of players is consistently closer to the ground truth for our method. Our method captures the multimodal distribution of player poses and generates vivid limb movements, while the baselines produce poses as the average of the distribution, resulting in reduced limb movement and tilted root joints. Additional samples are shown in \apref{ap:animation_evaluation}.}
  \label{fig:animation_ablation_qualitatives_tennis}
\end{figure}

\subsubsection{Comparison to Baselines}
\label{sec:animation_baselines}
Similarly to the synthesis model, we compare our animation model against the one of Playable Environments (PE) \cite{Menapace2022PlayableEnvironments}, the most related to our work since it operates on a similar environment representation. While the baseline jointly learns discrete actions and generates sequences conditioned on such actions, we assume the text action representations to be available in our task, so, for fairness of evaluation, we introduce our same text encoder $\nettext$ in the baseline to make use of the action information. To reduce computation, we perform the comparison using half of the computational resources and a reduced training schedule, consequently, we also retrain our model, producing a reduced variant (Ours Small). To render results we always make use of our synthesis model.

We show results averaged over all inference tasks in Tab.~\ref{table:animation_merged} and report the results for each task in \apref{ap:animation_evaluation}. Our method outperforms the baseline in all evaluation tasks according to both L2 and FD metrics. From the qualitative results in Fig.~\ref{fig:animation_ablation_qualitatives_tennis} and in accordance with the FD metrics, we notice that our method produces more realistic player poses with respect to PE that tends to keep player poses close to the average pose and to slide the players on the scene. We attribute this difference to the use of the diffusion framework in our method. Consider the example of generating a player walking forward. It is equally probable that the player moves the left or right leg first. In the case of a reconstruction-based training objective such as the main one of PE, the model is encouraged to produce an average leg movement result that consists in not moving the legs at all. On the other hand, diffusion models learn the multimodal distributions of the motion, thus they are able to sample one of the possible motions without averaging its predictions.

\begin{figure*}
\includegraphics[width=\textwidth]{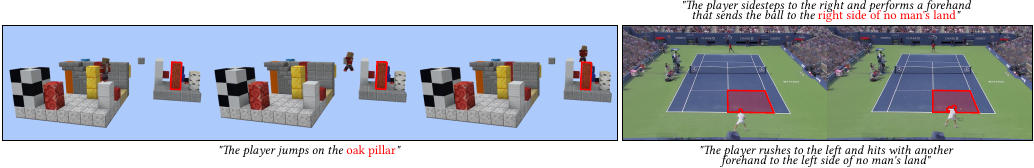}
  \caption{Behavior of the model when implausible actions are provided. In the left example, the model generates an irrealistic long jump to reach the specified pillar. In the right example, the bottom player is instructed to move left to intercept a ball coming to his right. In this case, the left movement command is ignored by the model to produce the closest plausible outcome.}
  \label{fig:impossible_actions}
\end{figure*}

\begin{table}[t]
\caption{Animation model comparison with baselines and ablation with results averaged over all inference tasks. Position and Joints 3D in meters, Root angle in axis-angle representation.}

\label{table:animation_merged}

\begin{center}

\footnotesize
\begin{tabular}{lcccccc}
\toprule
\multirow{2}{*}{\emph{Tennis}}  & \multicolumn{2}{c}{\emph{Position}} & \multicolumn{2}{c}{\emph{Root angle}} & \multicolumn{2}{c}{\emph{Joints 3D}} \\
 & L2$\downarrow$  & FD$\downarrow$ & L2$\downarrow$  & FD$\downarrow$ &  L2$\downarrow$  & FD$\downarrow$ \\

\midrule
PE & {3.291} & {229.112} & {1.126} & {15.953} & {0.303} & {53.242} \\
Rec. LSTM & {1.597} & {7.253} & \cellthird{0.907} & {7.051} & {0.193} & {16.735} \\
Rec. Transf. & \cellfirst{1.074} & \cellthird{4.402} & \cellfirst{0.767} & \cellthird{6.838} & \cellthird{0.175} & \cellthird{14.845} \\
Ours Small & \cellthird{1.380} & \cellsecond{1.443} & {1.014} & \cellsecond{0.560} & \cellsecond{0.148} & \cellsecond{1.253} \\
Ours & \cellsecond{1.099} & \cellfirst{0.929} & \cellsecond{0.844} & \cellfirst{0.356} & \cellfirst{0.129} & \cellfirst{0.836} \\

\midrule
\multirow{2}{*}{\emph{Minecraft}}  & \multicolumn{2}{c}{\emph{Position}} & \multicolumn{2}{c}{\emph{Root angle}} & \multicolumn{2}{c}{\emph{Joints 3D}} \\
 & L2$\downarrow$  & FD$\downarrow$ & L2$\downarrow$  & FD$\downarrow$ &  L2$\downarrow$  & FD$\downarrow$ \\
\midrule
PE & {2.739} & {105.973} & {1.620} & \cellthird{31.232} & \cellthird{0.311} & \cellthird{39.572} \\
Rec. LSTM & {2.292} & \cellthird{47.296} & {1.702} & {49.971} & {0.489} & {99.843} \\
Rec. Transf. & \cellthird{2.154} & {53.198} & \cellthird{1.430} & {36.123} & {0.385} & {69.977} \\
Ours Small & \cellsecond{1.084} & \cellfirst{4.461} & \cellsecond{1.077} & \cellsecond{6.016} & \cellsecond{0.140} & \cellsecond{3.590} \\
Ours & \cellfirst{1.065} & \cellsecond{4.815} & \cellfirst{0.956} & \cellfirst{4.083} & \cellfirst{0.132} & \cellfirst{3.360} \\

\bottomrule
\end{tabular}
\end{center}
\end{table}

\subsubsection{Ablation}
\label{sec:animation_model_ablation}
To validate this hypothesis and demonstrate the benefits of our diffusion formulation, we produce two variations of our method. The first substitutes the diffusion framework with a reconstruction objective, keeping the transformer-based architecture unaltered. The second in addition to using the reconstruction objective models $\netanimation$ using an LSTM, similarly to the PE baseline. Differently from the PE baseline, however, this variant does not make use of adversarial training and employs a single LSTM model for all objects, rather than a separate model for each.

We show results in Tab.~\ref{table:animation_merged}. Our model consistently outperforms the baselines in terms of FD, showing a better ability to capture realistic sequences. Consistently with our assessment in Sec.~\ref{sec:animation_baselines}, 
Fig.~\ref{fig:animation_ablation_qualitatives_tennis} shows that our method trained with a reconstruction objective produces player movement with noticeable artifacts analogously to PE, validating the choice of the diffusion framework.

\subsection{Limitations}
\label{ap:limitations}

Since the model is trained on a dataset showing only plausible actions, the model's behavior is not defined when an \emph{implausible} action is specified, such as hitting a ball while moving in the wrong direction to intercept it or jumping on a pillar that is out of reach. In these cases, we find the model to ignore the implausible part of the command and produce the closest plausible command or, less frequently, to produce implausible outcomes such as irrealistic long jumps (see Fig.~\ref{fig:impossible_actions}). In addition, the model does not generate actions extremely out of distribution such as performing a backflip or doing a push-up. \change{This aspect could be addressed by jointly training the animation model on multiple diverse datasets, which we consider an interesting future direction.}

While our Tennis dataset contains varied text annotations that allow the model to generalize to text inputs with varied structure, our Minecraft dataset's synthetic text annotations are less varied and the fixed synthetic structure of sentences tends to be memorized, making the model less effective if a different syntax is used (see Sec.~\ref{ap:language_robustness}). To address this issue, a more sophisticated algorithm can be employed to generate action annotation on the Minecraft dataset.

Our model learns to associate referential language to scene coordinates rather than the appearance of the referred object, and the model memorizes the position of contact surfaces. While tennis scenes always have the same structure, for Minecraft the model cannot generalize to different scenes. This concern can be addressed by conditioning the animation model on the scene's geometry, which we leave as future work.

\change{We find our animation model to overfit to the Tennis dataset when less than 60\% of the training data is used (see \apref{ap:animation_model_dataset_size_ablation}). We leave as an interesting avenue of future work the investigation of regularization techniques such as dropout or weight decay that have the potential to reduce overfitting in this scenario.}

Our animation model outperforms baselines that operate under the same data assumptions \cite{Menapace2022PlayableEnvironments} in terms of animation quality. With respect to recent character animation methods \cite{starke2019neural,starke2020local,holden2020learned} making use of richly annotated motion capture data and dataset-specific handcrafted optimizations (see Sec.~\ref{sec:character_animation}), our method demonstrates more advanced game dynamics and game AI modeling capabilities, but produces foot sliding artifacts. We expect continuous improvements in diffusion models to alleviate such artifacts and expect further improvements by considering different parametrizations of pose parameters taking into consideration the distance of limbs from the terrain, which we will explore in future work.

Lastly, our animation model does not yet produce results in real-time. We discuss inference speed and strategies to make the model real-time in \apref{ap:inference_speed}. Improving the sampling speed of diffusion models is an actively investigated problem \cite{salimans2022progressive,meng2022on,song2021denoising} that is orthogonal to ours.

\section{Conclusions}

In this paper, we demonstrate the feasibility of learning \change{game models able to answer challenging user prompts} and show that textual action representations are critical for unlocking fine-grained control over the generation process, and enabling compelling constraint- and goal-driven generation applications. 
These results, jointly with two richly-annotated text-video datasets, pave the way towards learning game models for complex, real-world scenes.

\section{Acknowledgements}
We would like to thank Christian Theobalt for his feedback on the manuscript draft, Denys Poluyanov, Eugene Shevchuk and Oleksandr Pyshchenko for the useful discussion and validation of the use cases of PGMs, Maryna Diakonova for her support in data labeling, and Anton Kuzmenko and Vadym Hrebennyk for their assistance in creating the accompanying video.

This work was partially supported by the EU HEU AI4TRUST (101070190) project.

\bibliographystyle{ACM-Reference-Format}
\bibliography{bibliography}

\clearpage
\appendix
\section{Datasets}
\label{ap:datasets}

In this section, we give additional details on the dataset, including the dataset collection process, Minecraft 3D skeleton format, additional dataset statistics, and dataset samples.

\subsection{Tennis Dataset Collection}
\label{ap:tennis_dataset_collection}
We build our tennis dataset starting from the one of \cite{Menapace2022PlayableEnvironments}. However, we notice that such dataset has an imprecise camera calibration and lacks information such as 3D player poses and 3D localization of the ball. Thus, we only retain the original videos and acquire new annotations. We describe the process in the following sections and release all code related to dataset creation.

\subsubsection{Camera Calibration}
To improve camera calibration, we notice that the original dataset bases its camera calibration on field keypoints detected using \cite{farin2003robustcameracalibration}, but such keypoint estimates are noisy. To overcome this issue, we manually annotate a subset of 10569 frames with field keypoint information and train a keypoint detection model inspired by ICNet \cite{zhao2018icnet}, which we choose due to its reduced memory footprint which allows us to train the model in full 1920x1080px resolution for best results. The detected keypoints are filtered and used to produce camera calibration. Compared to the camera calibration of \cite{Menapace2022PlayableEnvironments}, we notice less jitter and are able to successfully perform camera calibration on a larger number of video instances.

\subsubsection{3D Ball Localization}
To produce 3D ball localization, we first build a 2D ball detector following the same approach used for field keypoints localization, starting from 17330 manually annotated frames. In addition to 2D ball localization, we manually annotate the projection of the ball on the field plane for a set of keyframes defined as the frames where contact between the ball and an object different than the field happens or the first and last frames of the video with a visible ball. The field plane projections of the ball in conjunction with the camera calibration results and 2D ball detections can be used to recover the 3D ball position in those frames. 

We assume that between the keyframes, no contact happens that significantly alters the horizontal speed of the ball apart from air drag. In practice, contact between the ball and the field during bounces does affect ball speed, and we take account of it in a second, refinement phase. We thus model the horizontal ball position on the line between the ball positions at two consecutive keyframes by solving the linear motion equation under air drag:
\begin{equation}
\label{eq:drag_motion}
\vecpoint(t) = \vecpoint_0 \frac{\log(1 + \airdrag \vecvelocity_0 t)}{\airdrag},
\end{equation}
where $\vecpoint_0$ is the initial position, $\vecvelocity_0$ is the initial velocity, $t$ is time and $\airdrag$ is an estimated coefficient summarizing fluid viscosity, drag coefficient, and shape of the ball. Note that the effects of gravity are ignored in the equation.
$\airdrag$ can be estimated by inverting Eq.~\eqref{eq:drag_motion}, based on initial ball speed measurements for $\vecvelocity_0$ that can be extracted from the videos thanks to the service ball speed radars installed on tennis fields, and the positions the ball at keyframes. Given the ball's horizontal position on the line joining the 3D ball position at the preceding and succeeding keyframes, we can recover its 3D position by intersecting the camera ray passing from the 2D projection of the ball on that frame with the plane parallel to the net that intersects with the ball's horizontal position.

To improve the precision of results and account for horizontal ball speed changes during bounces, in a second phase we detect bounces between the ball and the field and impose that the ball touches the field at those positions, by considering them as additional keyframes and repeating the procedure. Finally, to calibrate frames with missing 2D ball detections (eg. ball thrown high above the camera frames or heavy blur and image compression artifacts), we recover the ball position by fitting a ballistic trajectory using 3D ball localization from neighboring frames.

\subsubsection{3D Player Poses}
To recover 3D player poses, we rely on the 3DCrowdNet pose estimator \cite{choi2022learning} which we find robust to the presence of frequent overlaps between players and referees, player limbs blur, and low player resolution. 3DCrowdNet assumes 2D joint locations to be given as input, so we produce them using the state-of-the-art 2D pose estimator VitPose \cite{xu2022vitpose} which we find robust to blur, reduced player size, and occlusions. The extracted 3D skeletons however are expressed under the coordinate system of a framework-predicted camera. We make use of a PnP \cite{EPnP} procedure to register the 3D skeletons to our calibrated camera and reduce depth estimation errors by placing the estimated 3D skeletons with their feet touching the ground. Note that, while 3DCrowdNet regresses full SMPL \cite{loper2015smpl} parameters and meshes, we only make use of 3D joint locations and joint angles. SMPL body shape parameters are nevertheless included in the dataset to support its different use cases.

\subsubsection{Text Action Annotation}
We manually annotate each video sequence using a text caption for each player and frame. Each caption focuses on the action being performed by the player in that instant and captures several aspects of the action. The caption captures where the player is moving and how the player is moving, i.e. the player is running, walking, sliding, or falling, the player is moving to its left, towards the net, across the baseline. When a player is performing a ball-hitting action, the particular type of tennis shot being performed is presented, e.g. a smash, a serve, a lob, a backhand, a volley, and the location where the ball is aimed is described. We report text annotation statistics in Tab.~\ref{table:dataset_statistics}.

\subsubsection{UI Elements Annotation}
We manually annotate each video sequence with a set of 2D bounding boxes indicating the places where 2D UI elements such as scoreboards or tournament logos may appear during the sequence.

\subsection{Minecraft Skeleton Format}
\label{ap:minecraft_skeleton_format}
We adopt a skeletal player representation that divides the Minecraft body into 6 parts: head, torso, left and right arm, and left and right leg. We place 6 corresponding joints at the bottom of the head, top of the torso, shoulders, and top of the legs. Following the internal Minecraft skeletal representation, a root joint is added that is the parent of the 6 joints. We extend this representation by introducing 6 additional joints at the top of the head, top of the torso, bottom of the arms, and bottom of the legs. The additional joints have as parents the original joint positioned on the same body part. While the additional 6 joints are always associated with a zero rotation, we find their introduction convenient for skeleton visualization purposes. Fig.~\ref{fig:dataset_samples} provides a visualization of such skeletons.

\subsection{Additional Dataset Statistics}
\begin{figure}
\centering
\begin{subfigure}{0.45\columnwidth}
    \includegraphics[width=\columnwidth]{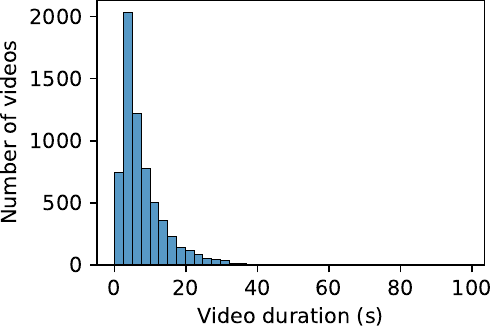}
    \caption{Distribution of video durations in the Tennis dataset.}
    \label{fig:tennis_video_duration}
\end{subfigure}
\hfill
\begin{subfigure}{0.45\columnwidth}
    \includegraphics[width=\columnwidth]{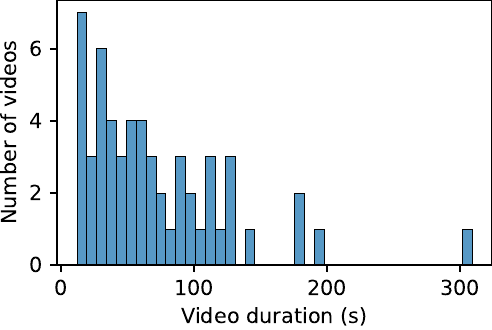}
    \caption{Distribution of video durations in the Minecraft dataset.}
    \label{fig:minecraft_video_duration}
\end{subfigure}
\hfill
\begin{subfigure}{0.45\columnwidth}
    \includegraphics[width=\columnwidth]{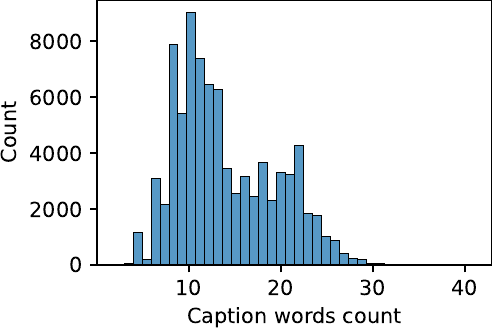}
    \caption{Distribution of words per caption in the Tennis dataset.}
    \label{fig:tennis_captions_length}
\end{subfigure}
\hfill
\begin{subfigure}{0.45\columnwidth}
    \includegraphics[width=\columnwidth]{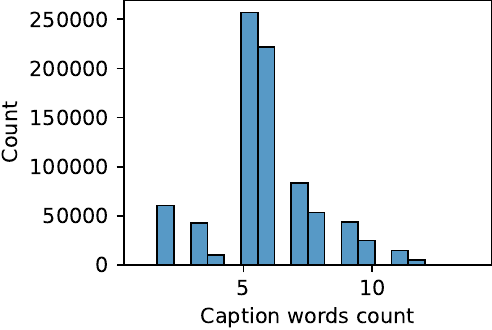}
    \caption{Distribution of words per caption in the Minecraft dataset.}
    \label{fig:minecraft_captions_length}
\end{subfigure}
        
\caption{Dataset statistics for the Tennis and Minecraft datasets.}
\label{fig:dataset_statistics}
\end{figure}

We provide the main dataset statistics in Tab.~\ref{table:dataset_statistics}, with additional ones in Fig.~\ref{fig:dataset_statistics}, where we plot the distribution of video lengths in the dataset and the average number of words in each caption. The Tennis dataset features manually-annotated captions which contain a greater number of words with respect to the synthetic annotations in the Minecraft dataset.

\subsection{Dataset Samples}
\label{ap:dataset_samples}

\begin{figure*}
\includegraphics{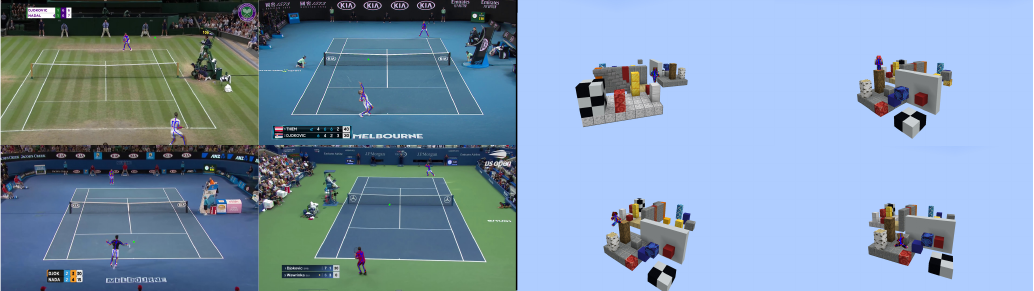}
  \caption{Sampled frames from the Tennis (left) and Minecraft (right) datasets. 3D skeletons are annotated in blue, while the 3D ball is visualized in green.}
  \label{fig:dataset_samples}
\end{figure*}

We show samples from the Minecraft and Tennis dataset in Fig.~\ref{fig:dataset_samples}.

We now show a non-curated set of captions extracted from the Tennis dataset:
\begin{itemize}
\item \emph{``the player prepares to hit the ball but stops, opponent hits the net''}
\item \emph{``the player starts to move to the left  when the opponent sends the ball out of the field''}
\item \emph{``the player moves diagonally to the right and forward to the right side of the baseline and sends the ball to the right side of no man's land with a forehand}
\item \emph{``the player takes sidestep to the right and hits the ball with a backhand that sends the ball to the right side of the no man's land}
\item \emph{``the player moves left to hit the ball but stops halfway}
\item \emph{``the player sidesteps to the left and stops, because the ball goes out of bounds}
\end{itemize}

We report a set of peculiar words extracted from the set of words with the lowest frequency on the Tennis dataset:
\emph{``scratching''},
\emph{``inertia''},
\emph{``previously''},
\emph{``realize''},
\emph{``understands''},
\emph{``succeed''},
\emph{``bind''},
\emph{``touched''},
\emph{``circling''},
\emph{``approaching''},
\emph{``bolting''},
\emph{``entering''},
\emph{``ducks''},
\emph{``reaction''},
\emph{``repeat''},
\emph{``wipes''},
\emph{``abruptly''},
\emph{``preparation''},
\emph{``dramatic''},
\emph{``soft''},
\emph{``celebrating''},
\emph{``losing''},
\emph{``strides''},
\emph{``dart''},
\emph{``reacts''},
\emph{``block''},
\emph{``sideway''},
\emph{``ending''},
\emph{``becomes''},
\emph{``dismissively''},
\emph{``continuous''},
\emph{``squat''},
\emph{``says''},
\emph{``intends''},
\emph{``ricochet''},
\emph{``delays''},
\emph{``night''},
\emph{``guess''},
\emph{``manage''},
\emph{``already''},
\emph{``correctly''},
\emph{``anticipation''},
\emph{``unsuccessfully''},
\emph{``inaccurate''},
\emph{``deflection''},
\emph{``properly''},
\emph{``swinging''}.

We show a non-curated set of captions extracted from the Minecraft dataset:
\begin{itemize}
    \item \emph{``the player falls on the birch pillar''}
    \item \emph{``the player moves fast north, jumps''}
    \item \emph{``the player jumps on the intermediate wooden pillar''}
    \item \emph{``the player falls on the platform opposite to the stairs''}
    \item \emph{``the player runs to the big stone platform''}
    \item \emph{``the player climbs down and does not rotate''}
    \item \emph{``the player moves south east, jumps and rotates counterclockwise''}
    \item \emph{``the player runs to the red decorated block''}
\end{itemize}

We list a set of peculiar words from the Minecraft dataset:
\emph{``nothing''},
\emph{``facing''},
\emph{``space''},
\emph{``level''},
\emph{``map''},
\emph{``leading''},
\emph{``opposite''},
\emph{``edge''}.

\begin{table}[t]
\caption{Dataset statistics for the Tennis and Minecraft datasets.}

\label{table:dataset_statistics}

\begin{center}

\footnotesize
\begin{tabular}{lcc}
\toprule
 & Tennis  & Minecraft \\
\midrule

Sequences: & 7112 & 61 \\
\;\;\;\emph{train} & 5690 & 51 \\
\;\;\;\emph{validation} & 711 & 5 \\
\;\;\;\emph{test} & 711 & 5 \\
\midrule
Duration: & 15.5h & 1.21h \\
\;\;\;\emph{train} & 12.4h & 0.952h \\
\;\;\;\emph{validation} & 1.59h & 0.16h \\
\;\;\;\emph{test} & 1.52h & 0.101 \\
\midrule
Annotated frames: & 1.12M & 68.5k \\
\;\;\;\emph{train} & 1.05M & 64.5k \\
\;\;\;\emph{validation} & 135k & 11.2k \\
\;\;\;\emph{test} & 130k & 7.06k \\
\midrule
Resolution & 1920x1080px & 1024x576px \\
Framerate & 25fps & 20fps \\
\midrule
Captions & 84.1k & 818k \\
\;\;\;\emph{of which unique} & 25.5k & 1.24k \\
Unique words & 915 & 117 \\
Avg. words & 13.8 & 5.85 \\
Avg. span & 1.32s & 0.500s \\
Parts of sentence: & & \\
\;\;\;\emph{Nouns} & 32.3\% & 36.2\% \\
\;\;\;\emph{Verbs} & 11.9\% & 17.4\% \\
\;\;\;\emph{Adjectives} & 3.08\% & 6.48\% \\
\;\;\;\emph{Adverbs} & 2.70\% & 11.7\% \\
\;\;\;\emph{Pronouns} & 0.18\% & 0.00\% \\
\;\;\;\emph{Articles} & 26.4\% & 8.03\% \\
\;\;\;\emph{Prepositions} & 7.89\% & 6.98\% \\
\;\;\;\emph{Numerals} & 0.11\% & 0.03\% \\
\;\;\;\emph{Particles} & 9.28\% & 1.50\% \\
\;\;\;\emph{Punctuation} & 1.76\% & 1.12\% \\
\;\;\;\emph{Others} & 0.00\% & 0.00\% \\
\bottomrule
\end{tabular}
\end{center}

\end{table}

\section{Object-Specific Synthesis Model Techniques}
\label{ap:object_specific_techniques}
The compositional nature of the synthesis module makes it possible to adopt object-specific techniques to model particular objects. In the following, we describe the techniques adopted to model balls (\apref{ap:ball_modeling}), rackets (\apref{ap:racket_modeling}), 2D UI elements (\apref{ap:ui_elements}), and skyboxes (\apref{ap:skybox_modeling}).

\subsection{Ball Modeling}
\label{ap:ball_modeling}

Fast-moving objects may appear blurred in real video sequences. This effect is frequent in ball objects found in sports videos and is thus desirable to model this effect. To model them, we adopt a procedure inspired by \cite{cook1984distributed}, which distributes multiple rays in time to model blur effects.
We extend the object properties of the ball object to also include a velocity vector $\vecvelocity$. Given the ball radius $r$ and an estimate for the shutter speed $t_c$ of the camera, we can compute in closed form the probability $p$ that a given point in space intersects with the ball object while the ball moves during the time the camera shutter remains open to capture the current frame. To model blur, we assign to each point a fixed density multiplied by $p$. Modeling $p$ in closed form avoids the need to sample multiple rays in time, improving performance.

\begin{figure} 
\includegraphics{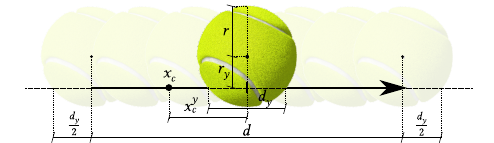}
  \caption{Visualization of the quantities involved in the computation of the probability $p$ that the ball intersects with a certain point in space during a randomly sampled time instant in the interval from the opening to the closing of the shutter of the camera to capture the current frame. The leftmost and rightmost balls depict the ball position at the times the camera shutter opens and closes respectively. For simplicity, we represent the space where the velocity vector of the ball is aligned with the $y$-axis.}
  \label{fig:ball_blur_diagram} 
\end{figure}

To compute $p$ (see Fig.~\ref{fig:ball_blur_diagram}), we first use the velocity vector $\vecvelocity$ to estimate the rotation $\tensrotation_b$ that maps each point $\vecpointbbox$ in the ball bounding box to a canonical space $\vecpointcanon$ in which the ball velocity vector is aligned to the positive $y$-axis $\vecpointcanon = \tensrotation_b \vecpointbbox$. 
Then we compute the distance traveled by the ball while the shutter remains open $d=||\vecvelocity||_2 t_c$. 
We then compute the useful cross-section of the ball $d_y$ that can intersect with $\vecpointcanon$ as the diameter of the circumference originating from the intersection between the ball and a plane with a distance from the ball center $r_y$ equal to the distance of $\vecpointcanon$ from the $y$-axis:
\begin{equation}
    d_y=
    \begin{cases}
    2r \sin \left(\arccos\left(\frac{r_y}{r}\right)\right) & \text{if } r_y \le r\\
    0              & \text{otherwise}
\end{cases}.
\end{equation}

Finally, $p$ equals the probability that an interval with size equal to the cross-section, positioned in a random portion of space contained inside an interval of size $d + d_y$, that represents the length of the space that has been touched by the ball while the shutter stays open, contains our point $\vecpointcanon$:
\begin{equation}
p(\vecpointcanon) = \max\left(0, \min\left(\min\left(\frac{d_y}{d}, 1\right), \frac{1}{2} + \frac{d_y}{2d} - \frac{|\vecpointcanon^y|}{d}\right)\right),
\end{equation}
where $\vecpointcanon^y$ is the $y$-axis coordinate of $\vecpointcanon$.

\subsection{Racket Modeling}
\label{ap:racket_modeling}

\begin{figure} 
\includegraphics{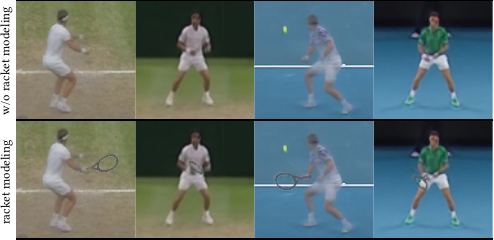}
  \caption{Examples of tennis scenes with and without inserted rackets.}
  \label{fig:racket_modeling} 
\end{figure} 

Modeling the scene as a composition of neural radiance fields allows applications such as the insertion of user-defined watertight 3D meshes into the scene. To do so, we first define the 3D bounding box for the mesh. Then, we extract the signed distance function (SDF) of the 3D mesh. To allow fast retrieval of SDF values during rendering, we sample SDF values along an enclosing voxel grid so that subsequently they can be efficiently retrieved using trilinear sampling. During neural rendering, when a sampled point intersects with the object's bounding box, we query its SDF function and assign a fixed, high density to points that fall inside the object. For simplicity, we assume the object has a uniform appearance and assign a fixed feature vector to such points. To attach the mesh to an articulated object, we align it to its desired position in the object's canonical space, select which joint the mesh should move according to, and we modify blending weights $\tensblendweights$ for the desired joint to have a high value in the region corresponding to the mesh (see Eq.~\eqref{eq:invlbs}).

We employ this technique on the Tennis dataset to manually insert rackets in the scene that cannot be easily learned since they appear frequently blurred and have no ground truth pose available. After the synthesis model is trained, we use this technique to insert a racket mesh in the hand of each player and configure it to move it according to the elbow joint. Fig.~\ref{fig:racket_modeling} shows examples of rackets inserted in tennis scenes.

When inserting additional objects at inference time, we find the enhancer model $\netenhancer$ may introduce artifacts at the contours of the inserted object. For this reason, we modify $\netenhancer$ with a masking mechanism that directly uses values from the NeRF-rendered RGB image $\tensimagergb$ before the enhancer rather than the enhanced image $\tensimagerec$ for pixels corresponding to the inserted object and its contour region.

\subsection{2D UI Elements}
\label{ap:ui_elements}

The presence of 2D user interfaces, such as scoreboards, in the training frames may cause artifacts in the final outputs due to attempts of the synthesis model to model these view-inconsistent elements \cite{Menapace2022PlayableEnvironments}. To address this issue, we assume that the potential regions where such interfaces may be present are known and we never sample training patches that intersect with these regions. In this way, the model does not attempt to generate such UI elements and instead models the underlying portion of the 3D scene using data from different views.

\subsection{Skybox Modeling}
\label{ap:skybox_modeling}
The Minecraft background is represented as a skybox that is modeled by extending the planar object modeling mechanism of Sec.~\ref{sec:canonical_volume}. In more detail, we sample the feature plane $\tensplane$ according to the ray's yaw and pitch of the current ray, which can be interpreted as querying points on the surface of a sphere with a radius approaching infinity.

\section{Deformation Modeling}
\label{ap:deformation_modeling}
In this section, we present additional details on the deformation model $\netdeformation$ used to render articulated objects such as humans. Given an articulated object, we assume its kinematic tree is known and that the transformation $[\tensrotation_j|\vectranslation_j]$ from each joint $j \in 1,...,\numjoints$ to the parent joint is part of the object's properties. From these we can follow the kinematic tree to derive transformations $[\tensrotation'_j|\vectranslation'_j]$ for each joint from the bounding box coordinate system to the canonical coordinate system. Intuitively, these transformations represent how to map a point $\vecpointbbox$ in the bounding box coordinate system belonging to the joint $j$ to the corresponding point $\vecpointcanon$ in the canonical space.

We implement a deformation procedure based on linear blend skinning (LBS) \cite{lewis2000pose} that establishes correspondences between points in the canonical space $\vecpointcanon$ and in the deformed bounding box space $\vecpointbbox$ by introducing blending weights $\vecblendweights$ for each point in the canonical space. These weights can be interpreted as the degree to which that point moves according to the transformation associated with that joint.
\begin{equation}
\label{eq:lbs}
\vecpointbbox = \sum_{j=1}^{\numjoints} w_j(\vecpointcanon)\left(\tensrotation_j^{\prime-1}  \vecpointcanon - \tensrotation_j^{\prime-1} \vectranslation'_j\right).
\end{equation}

During volumetric rendering, however, we sample points $\vecpointbbox$ in the bounding box space and query the canonical volume in the corresponding canonical space point $\vecpointcanon$. Doing so requires solving Eq.~\eqref{eq:lbs} for $\vecpointcanon$, which is prohibitively expensive \cite{li2022tava}. Inspired by HumanNeRF \cite{weng2022humannerf}, instead of modeling LBS weights $\vecblendweights$, we introduce inverse linear blending weights $\vecblendweights^b$:
\begin{equation}
\label{eq:invlbs}
    \vecblendweights_j^b(\vecpointbbox) = \frac{\vecblendweights_j(\tensrotation'_j \vecpointbbox + \vectranslation'_j)}{\sum_{j=1}^{\numjoints} \vecblendweights_j(\tensrotation'_p \vecpointbbox + \vectranslation'_j)}.
\end{equation}
such that the canonical point can be approximated as:
\begin{equation}
\label{eq:lbs_hnerf}
\vecpointcanon = \sum_{j=1}^{\numjoints} \vecblendweights_j^b(\vecpointbbox)\left(\tensrotation'_j \vecpointbbox + \vectranslation'_j\right).
\end{equation}

We parametrize the function $\vecblendweights$ mapping spatial locations in the canonical space to blending weights as a neural network. Similarly to $\netcanonical$, we employ 3D convolutions to map a fixed noise volume $\tensblendweightssmall \in \mathbb{R}^{\numfieldfeatsmall \times \numblendweightsheightsmall \times \numblendweightswidthsmall \times 
\numblendweightsdepthsmall}$ to a volume of blending weights $\tensblendweights \in \mathbb{R}^{\numjoints + 1 \times \numblendweightsheight \times \numblendweightswidth \times 
\numblendweightsdepth}$, where each channel represents the blending weights for each part, with an extra weight modeling the background. The volume channels are normalized using softmax, so that they sum to one, and can efficiently be queried using trilinear sampling.
To facilitate convergence, we exploit the known kinematic tree to build a prior over the blending weights that increases blending weights in the area surrounding each limb \cite{weng2022humannerf}.
\section{Implementation Details}

\subsection{Synthesis Model}
\label{ap:implementation_details_synthesis}

We model Minecraft scenes considering as objects the player, the scene, and the background. 
To model Tennis scenes, we consider as separate objects the two players, the ball, the field plane, and the vertical backplate at the end of the field. Both players share the same canonical representation. Note that the field and backplate are modeled as planar objects due to the lack of camera translation on the tennis dataset, which does not make it possible to reconstruct the geometry of static objects \cite{Menapace2022PlayableEnvironments}.

For each ray, we uniformly sample 32 points for players, 16 for the ball, 48 for the Minecraft scene, and 1 for all remaining objects that are modeled as planes. We do not employ hierarchical sampling, which we empirically found not to improve results. A patch size of 180x180px and of 128x128px are employed respectively for the Tennis and Minecraft datasets.

We model the initial blocks of the style encoder $\netstyle$ as the first two residual blocks of a pretrained ResNet 18 \cite{he2016deep}. To prevent players from being modeled as part of the background, we always sample images in pairs from each video and randomly swap the style codes $\vecstyle$ of corresponding objects \cite{Menapace2022PlayableEnvironments}.

To represent the player canonical radiance fields, we use a voxel $\tensvoxel$ with $\numfieldfeat=64$ features and $\numvoxelheight=\numvoxelwidth=\numvoxeldepth=32$. Deformations are represented using blending weights $\tensblendweights$ with $\numblendweightsheight=\numblendweightswidth=\numblendweightsdepth=32$. For the Minecraft scene, the size of the voxel $\tensvoxel$ is increased to $\numvoxelheight=\numvoxelwidth=\numvoxeldepth=128$. The Minecraft skybox is represented with feature planes $\tensplane$ with $\numfieldfeat=64$ features and size $\numplaneheight=\numplanewidth=256$. Due to their increased complexity and variety of styles, in the Tennis dataset feature planes $\tensplane$ with $\numfieldfeat=512$ features are adopted. The MLPs performing stylization of the canonical field features are modeled using 2 layers with a hidden dimension of 64, with a final number of output features $\numfieldfeat=19$, where the first 3 channels represent radiance.

\subsection{Animation Model}
\label{ap:implementation_details_animation}

For the text encoder, we model $\nettextenc$ as a frozen T5-Large 
\cite{raffel2022exploring} model and $\nettextagg$ as a transformer encoder \cite{vaswani2017attention} with 4 layers, 8 heads, and a feature size of 1024. For each sequence, the output $\vecactionemb$ of $\nettext$ is the transformer encoder output corresponding to the position of the end-of-sentence token in the input sequence. We experimented with mean pooling and a learnable class token with comparable results. Consistent with \cite{saharia2022imagen}, we found alternative choices for $\nettextenc$ (T5-Small, T5-Base \cite{raffel2022exploring} and the CLIP text encoder \cite{radford2021clip}) to underperform T5-Large.

For the temporal model $\netanimation$, we employ a transformer encoder with 12 layers, 12 heads, and 768 features.
To favor generalization to sequences of different lengths at inference time, we adopt relative positional encodings \cite{shaw2018self} that specify positional encodings based on the relative distance in the sequence between sequence elements.
We produce embeddings for the diffusion timestep $\difftimestep$ and framerate $\framerate$ using sinusoidal position encodings \cite{vaswani2017attention}.
Additionally, to enable the model to better distinguish between noisy sequence entries and conditioning signals, we find it beneficial to condition also on $\vecmask$ and $\vecmaskaction$ using the same weight demodulation layer.

 The temporal model receives a flattened sequence of object properties grouped and encoded as follows: the position of objects as the bounding box center point; the player poses expressed with joint translations and rotations separately, with rotations expressed in axis-angle representation, which we find to produce more realistic animations with respect to the 6D representation of \cite{zhou2019cvpr}; the ball speed vector expressed as its orientation in axis-angle representation and norm. Separating positions from joint translations and rotations has the practical implication that these properties can be independently used as conditioning signals during inference. This enables applications such as generating realistic joint rotations and translations given a sequence of object positions in time describing the object movement trajectory. We assume style to remain constant in the sequence, thus we do not include it as input to the model.

\section{Training Details}
\label{ap:training_details}
\change{In this section, we discuss training details for our synthesis model (Sec.~\ref{ap:training_details_synthesis}) and animation model (Sec.~\ref{ap:training_details_animation}). In addition, we provide training cost estimates\footnote{\label{lambda_cost}Cost estimate from \url{lambdalabs.com} GPU Cloud service} for the different model variants.}

\subsection{Synthesis Model}
\label{ap:training_details_synthesis}
We employ a reduced learning rate of $1e-5$ for the 3D CNNs that model the canonical radiance field voxels $\tensvoxel$ and blending weights $\tensblendweights$ that we find important to improve the learned geometry and avoid artifacts such as holes.

We train our full model on 8 A100 40GB GPUs for 4 days \change{(844\$)} and 2 days \change{(422\$)} respectively on the tennis and Minecraft datasets. We train the reduced version of our model (Ours Small) on 4 A100 40GB GPUs for 3 days \change{(317\$)} and 2 days \change{(211\$)} respectively for the Tennis and Minecraft datasets.

\subsection{Animation Model}
\label{ap:training_details_animation}
We create masks $\vecmask$ and $\vecmaskaction$ by randomly selecting one of the following masking strategies:
\begin{enumerate}[label=\roman*]
\item randomly mask each sequence element with a probability 0.25
\item randomly mask each sequence element with a probability 0.5
\item mask all sequence elements corresponding to a block of consecutive timesteps of random length
\item the complement of (iii)
\item mask all sequence elements corresponding to the last timesteps of the sequence
\item mask all sequence elements corresponding to a randomly chosen set of object properties
\end{enumerate}

With probability 0.5, we set $\vecmaskaction=1$, excluding actions from the masking operation, so that the model can learn to solve (ii), (iii), (iv) also in the scenario where text guidance is provided. We design the masking strategies to mimic masking configurations that are relevant to inference problems such as autoregressive generation (v), unconditional generation (vi), generating opponent responses to user actions (vi), sequence inpainting (iii), sequence outpainting (iv) and framerate increase (iii).

We train our full model on 8 A100 40GB GPUs for 15 days \change{(3168\$)} and 10 days \change{(2112\$)} respectively on the tennis and Minecraft datasets. We train the reduced version of our model (Ours Small) on 2 A100 40GB GPUs for 6 days \change{(317\$)} and 4 days \change{(211\$)} respectively for the Tennis and Minecraft datasets.
\section{Inference Details}
\label{ap:inference_details}

\subsection{Inference Speed}
\label{ap:inference_speed}
Our synthesis model renders images at 2.96fps over a single A100 GPU. We can parallelize inference by generating batches of 8 consecutive frames on separate GPUs for 23.7fps.
The animation model \change{has a throughput} of 1.08fps using 1000 diffusion sampling timesteps, measured by dividing the number of generated frames by the computation time at the end of the diffusion process. Meng \etal \cite{meng2022on} show that a reduction to 16 timesteps is possible with no or minimal loss in quality for a projected performance of 67.5fps. Hence, we believe our framework can be made real-time, which is a scope for future works. 

\subsection{Animation Model Inference Details}
\label{ap:animation_inference}

At inference time, the user is presented with a fully-masked, empty sequence $\vecsequencecond=0$, $\vecmask=0$, $\vectextactioncond=``"$, $\vecmaskaction=0$. Any object property can be specified as a conditioning signal in $\vecsequencecond$ and text action descriptions for any sequence timesteps can be provided in $\vectextactioncond$, with masks updated accordingly. The desired framerate $\framerate$ is also specified.

The text encoder $\nettext$ produces text embeddings $\vecactionemb$ as in Eq.~\eqref{eq:text_encoder}. Successively, the \emph{reverse process} is started at diffusion time $\difftimestep=\numdifftimesteps$, with $\vecsequenceprednoise_{\numdifftimesteps}$ sampled from the normal distribution. The DDPM sampler \cite{ho2020ddpm} queries the temporal model 
according to Eq.~\eqref{eq:temporal_model} to progressively denoise $\vecsequenceprednoise_\difftimestep$ and obtain the predicted sequence $\vecsequencepred=\vecsequenceprednoise_0$. The final sequence is obtained as $\vecsequence = \vecsequencepred + \vecsequencecond$, following Eq.~\eqref{eq:sequence_composition}.

\subsubsection{High Framerate Generation}
\label{ap:high_framerate_generation}
To produce sequences at the dataset framerate, we devise a two-stage sampling procedure designed to prevent an excessive increase in the sequence length. In the first stage, we sample the desired sequence at a low framerate $\framerate_1$. In the second stage, we exploit the masking mechanism and framerate conditioning to increase the framerate and, consequently, the length of the generated sequence. After the first stage, we consider a higher framerate $\framerate_2$ and extend the sampled sequence $\vecsequence$ with new states between existing ones, that we call keyframes, until the sequence length corresponding to $\framerate_2$ is reached. This sequence constitutes the new $\vecsequencecond$. Any previous action conditioning is copied in a new $\vectextactioncond$ in the corresponding keyframe locations. Masks are updated to be 1 in the position of the keyframes and 0 elsewhere. The sampling process is then repeated with the new conditioning signals and a sequence $\vecsequence$ is produced at the final framerate $\framerate_2$. To avoid an explosion in the length of the sequence, we exploit keyframes to divide the sequence into shorter chunks beginning and terminating at a keyframe, and sampling is performed separately on each chunk.

\subsubsection{Autoregressive Generation}
\label{ap:augoregressive_generation}

Our masking mechanism can be used to produce predictions autoregressively, enabling long sequence generation. Autoregressive generation can be obtained by considering a sequence $\vecsequencecond$ and removing the states corresponding to the first $t$ timesteps. $t$ timesteps are then added at the end of the sequence and a mask $\vecmask$ is created to zero out these additional $t$ steps. Conditioning signals can then be specified as desired for the last $t$ timesteps. When sampling $\vecsequencepred$, a prediction is produced for the additional timesteps and the procedure can be repeated.
\section{Applications}
\label{ap:applications}

\begin{figure*}
\includegraphics[width=\textwidth]{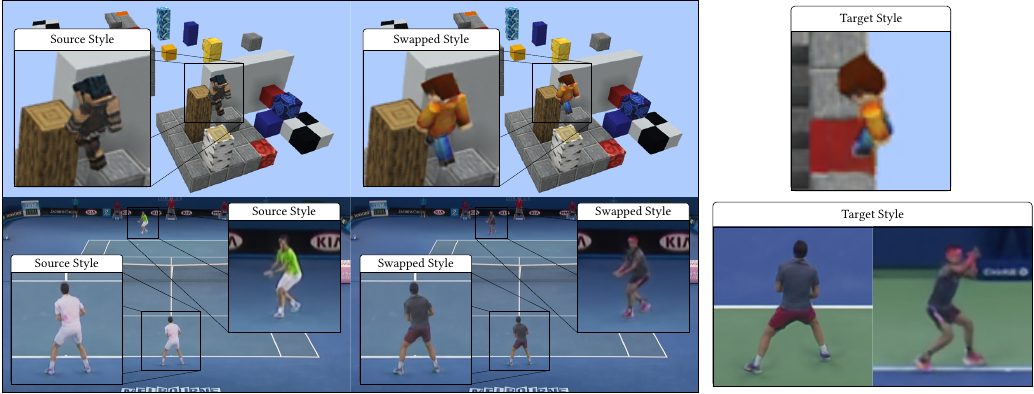}
  \caption{Style swap results on the Tennis and Minecraft datasets. We produce the central image by swapping the style code $\vecstyle$ for the players on the leftmost image with the ones from the rightmost image. Minecraft results are cropped for better visualization.}
  \label{fig:style_swap}
\end{figure*}

\begin{figure*}
\includegraphics[width=\textwidth]{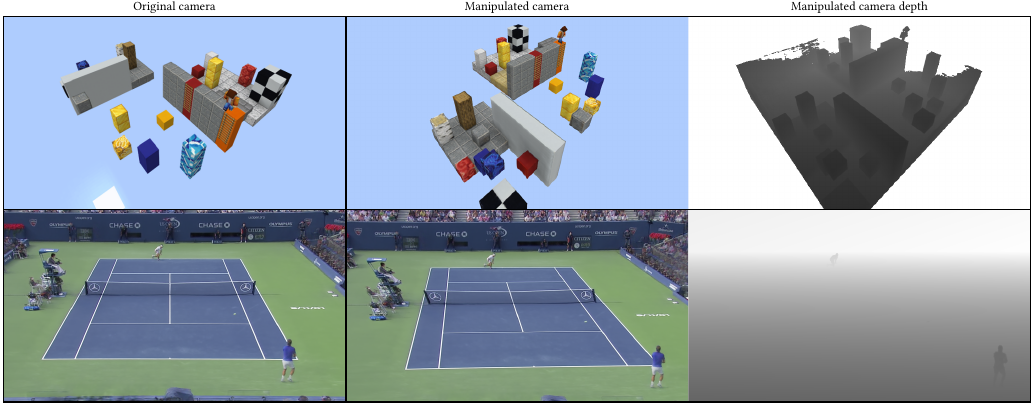}
  \caption{Camera manipulation results on the Tennis and Minecraft datasets. The lack of camera translation on the Tennis dataset does not allow to capture the 3D geometry of static objects, which is replaced by a prior \cite{Menapace2022PlayableEnvironments}. Minecraft results are cropped for better visualization.}
  \label{fig:camera_manipulation_qualitatives}
\end{figure*}

To demonstrate style swapping capabilities, in Fig.~\ref{fig:style_swap} we swap the style of the player $\vecstyle$ in the original image with the one from a target image. In addition, our synthesis model renders the current state of the environment from a user-defined perspective, similarly to the rendering component of a game engine. This enables our QGE to perform novel view synthesis as shown in Fig.~\ref{fig:camera_manipulation_qualitatives}.
\section{Additional Evaluation}
\label{ap:evaluation}

\subsection{Robustness to prompt variations}
\label{ap:language_robustness}
We perform a study on the Tennis dataset to evaluate the capability of our animation model to support diverse language prompts. We randomly sampled 50 prompts from our tennis dataset and asked \emph{ChatGPT to ``Produce a semantically equivalent reformulation of the prompt ‘<prompt>’''}. The sentence similarity between original and reformulated prompts measured by Jaccard similarity on their 3-grams is 0.390, while it is 0.203 for random dataset prompt pairs, indicating high diversity. As an example, the prompt \emph{``the player stops and quickly runs to the right and hits the ball with a backhand towards the center of no man's land''} is reformulated to \emph{``The player comes to a stop and rapidly sprints towards the right before executing a backhand stroke that directs the ball towards the center of no man's land''}, and prompt \emph{``the player prepares to hit the ball but stops, returning ball hits the net''} is transformed to \emph{``The player readies themselves to strike the ball, but abruptly halts and as a result, the returned ball collides with the net''}.

Successively, we run an AMT user study. Users are shown a video and two prompts (the true prompt used to produce the video, and a random negative prompt) and they are asked to recognize which of two prompts is the one used to produce a certain video. The average accuracy over 500 responses is $74.4\%$ and $77.1\%$ for videos produced using reformulated and dataset prompts respectively, indicating capability of the model to generate videos matching prompts independently from the form of the used language.

The model trained on Minecraft allows for limited prompt variation due to the synthetic nature of the training language whose limited variation does not enable the model to learn generalization capabilities to different sentence structures as the ones acquired for Tennis. This limitation could be addressed by improving the synthetic language generation process, which we leave as future work.

\subsection{Animation Model Evaluation}
\label{ap:animation_evaluation}
\begin{figure}
\includegraphics{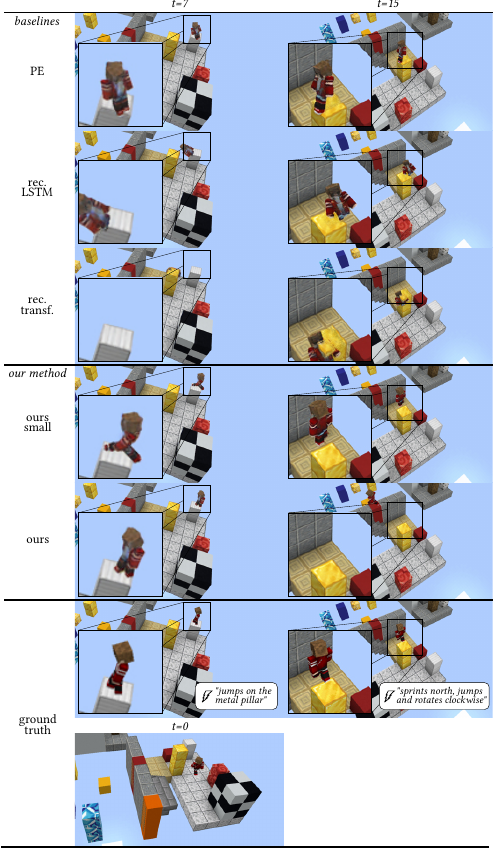}
  \caption{Qualitatives results on the Minecraft dataset. Sequences are produced in a video prediction setting that uses the first frame object properties and all actions as conditioning.}
  \label{fig:animation_ablation_qualitatives_minecraft}
\end{figure}

In Tab.~\ref{table:animation_tennis} and Tab.~\ref{table:animation_minecraft} we show evaluation results for each inference task respectively on the Tennis and Minecraft datasets. In Fig.~\ref{fig:animation_ablation_qualitatives_minecraft} we show qualitative results on the Minecraft dataset.

\begin{table}[t]
\caption{Animation model comparison with baselines and ablation on the Tennis dataset. Position and Joints 3D in meters, Root angle in axis-angle representation.}

\label{table:animation_tennis}

\begin{center}

\footnotesize
\begin{tabular}{lcccccc}
\toprule
\multicolumn{1}{c}{}  & \multicolumn{2}{c}{\emph{Position}} & \multicolumn{2}{c}{\emph{Root angle}} & \multicolumn{2}{c}{\emph{Joints 3D}} \\
 & L2$\downarrow$  & FD$\downarrow$ & L2$\downarrow$  & FD$\downarrow$ &  L2$\downarrow$  & FD$\downarrow$ \\

\midrule
& \multicolumn{6}{c}{\emph{Action conditioned video prediction}}\\
\midrule
PE & {3.117} & {87.688} & {1.182} & {12.627} & {0.277} & {30.711} \\
Rec. LSTM & {1.753} & {7.413} & \cellthird{1.100} & {8.416} & {0.234} & {18.455} \\
Rec. Transf. & \cellsecond{1.183} & \cellthird{2.996} & \cellfirst{0.913} & \cellthird{7.566} & \cellthird{0.212} & \cellthird{15.976} \\
Ours Small & \cellthird{1.244} & \cellsecond{1.071} & {1.187} & \cellsecond{0.601} & \cellsecond{0.178} & \cellsecond{1.570} \\
Ours & \cellfirst{1.064} & \cellfirst{0.846} & \cellsecond{0.961} & \cellfirst{0.421} & \cellfirst{0.153} & \cellfirst{1.049} \\
\midrule
& \multicolumn{6}{c}{\emph{Unconditional video prediction}}\\
\midrule
PE & {3.973} & {146.019} & {1.604} & {30.448} & {0.437} & {78.835} \\
Rec. LSTM & \cellthird{2.064} & {11.283} & \cellsecond{1.224} & \cellthird{14.860} & {0.264} & {28.736} \\
Rec. Transf. & \cellfirst{1.649} & \cellthird{10.514} & \cellfirst{1.123} & {15.648} & \cellthird{0.251} & \cellthird{27.258} \\
Ours Small & {2.352} & \cellsecond{2.271} & {1.455} & \cellsecond{0.781} & \cellsecond{0.213} & \cellsecond{1.827} \\
Ours & \cellsecond{1.925} & \cellfirst{1.377} & \cellthird{1.277} & \cellfirst{0.518} & \cellfirst{0.192} & \cellfirst{1.261} \\
\midrule
& \multicolumn{6}{c}{\emph{Opponent modeling}}\\
\midrule
PE & {4.353} & {641.976} & {0.903} & {13.955} & {0.251} & {62.981} \\
Rec. LSTM & {1.581} & {5.507} & \cellsecond{0.697} & {2.517} & {0.143} & \cellthird{10.443} \\
Rec. Transf. & \cellsecond{1.169} & \cellthird{3.735} & \cellfirst{0.631} & \cellthird{2.514} & \cellthird{0.138} & {10.519} \\
Ours Small & \cellthird{1.578} & \cellsecond{2.243} & {0.832} & \cellsecond{0.560} & \cellsecond{0.114} & \cellsecond{0.851} \\
Ours & \cellfirst{1.153} & \cellfirst{1.349} & \cellthird{0.703} & \cellfirst{0.288} & \cellfirst{0.101} & \cellfirst{0.558} \\
\midrule
& \multicolumn{6}{c}{\emph{Sequence completion}}\\
\midrule
PE & {1.720} & {40.766} & {0.814} & {6.783} & {0.246} & {40.441} \\
Rec. LSTM & {0.990} & {4.809} & {0.606} & {2.411} & {0.132} & {9.305} \\
Rec. Transf. & \cellsecond{0.294} & \cellthird{0.364} & \cellfirst{0.403} & \cellthird{1.623} & \cellthird{0.100} & \cellthird{5.628} \\
Ours Small & \cellthird{0.344} & \cellsecond{0.187} & \cellthird{0.581} & \cellsecond{0.301} & \cellsecond{0.088} & \cellsecond{0.765} \\
Ours & \cellfirst{0.252} & \cellfirst{0.143} & \cellsecond{0.437} & \cellfirst{0.198} & \cellfirst{0.069} & \cellfirst{0.478} \\
\midrule
& \multicolumn{6}{c}{\emph{Average}}\\
\midrule
PE & {3.291} & {229.112} & {1.126} & {15.953} & {0.303} & {53.242} \\
Rec. LSTM & {1.597} & {7.253} & \cellthird{0.907} & {7.051} & {0.193} & {16.735} \\
Rec. Transf. & \cellfirst{1.074} & \cellthird{4.402} & \cellfirst{0.767} & \cellthird{6.838} & \cellthird{0.175} & \cellthird{14.845} \\
Ours Small & \cellthird{1.380} & \cellsecond{1.443} & {1.014} & \cellsecond{0.560} & \cellsecond{0.148} & \cellsecond{1.253} \\
Ours & \cellsecond{1.099} & \cellfirst{0.929} & \cellsecond{0.844} & \cellfirst{0.356} & \cellfirst{0.129} & \cellfirst{0.836} \\

\bottomrule
\end{tabular}
\end{center}
\end{table}

\begin{table}[t]
\caption{Animation model comparison with baselines and ablation on the Minecraft dataset. Position and Joints 3D in meters, Root angle in axis-angle representation.}

\label{table:animation_minecraft}

\begin{center}

\footnotesize
\begin{tabular}{lcccccc}
\toprule
\multicolumn{1}{c}{}  & \multicolumn{2}{c}{\emph{Position}} & \multicolumn{2}{c}{\emph{Root angle}} & \multicolumn{2}{c}{\emph{Joints 3D}}  \\
 & L2$\downarrow$  & FD$\downarrow$ & L2$\downarrow$  & FD$\downarrow$ &  L2$\downarrow$  & FD$\downarrow$ \\

\midrule
& \multicolumn{6}{c}{\emph{Action conditioned video prediction}}\\
\midrule
PE & {2.720} & {90.904} & {1.822} & \cellthird{23.949} & \cellthird{0.365} & \cellthird{47.956} \\
Rec. LSTM & \cellthird{2.623} & \cellthird{54.927} & {2.040} & {62.363} & {0.579} & {118.592} \\
Rec. Transf. & {2.798} & {76.582} & \cellthird{1.794} & {52.677} & {0.506} & {100.731} \\
Ours Small & \cellsecond{0.533} & \cellfirst{2.494} & \cellsecond{0.901} & \cellsecond{5.624} & \cellsecond{0.145} & \cellsecond{4.083} \\
Ours & \cellfirst{0.523} & \cellsecond{2.582} & \cellfirst{0.749} & \cellfirst{4.578} & \cellfirst{0.135} & \cellfirst{3.794} \\
\midrule
& \multicolumn{6}{c}{\emph{Unconditional video prediction}}\\
\midrule
PE & {3.994} & {197.434} & {2.111} & {59.112} & \cellthird{0.370} & \cellthird{46.665} \\
Rec. LSTM & {2.850} & \cellthird{68.915} & {1.999} & {69.886} & {0.581} & {121.187} \\
Rec. Transf. & \cellthird{2.834} & {76.780} & \cellsecond{1.795} & \cellthird{50.871} & {0.480} & {87.584} \\
Ours Small & \cellsecond{2.341} & \cellfirst{9.795} & \cellthird{1.814} & \cellsecond{8.969} & \cellsecond{0.199} & \cellsecond{4.422} \\
Ours & \cellfirst{2.330} & \cellsecond{11.032} & \cellfirst{1.685} & \cellfirst{5.648} & \cellfirst{0.197} & \cellfirst{4.385} \\
\midrule
& \multicolumn{6}{c}{\emph{Sequence completion}}\\
\midrule
PE & {1.504} & {29.582} & {0.926} & {10.634} & {0.197} & {24.096} \\
Rec. LSTM & {1.401} & {18.044} & {1.066} & {17.664} & {0.309} & {59.750} \\
Rec. Transf. & \cellthird{0.830} & \cellthird{6.232} & \cellthird{0.700} & \cellthird{4.822} & \cellthird{0.170} & \cellthird{21.615} \\
Ours Small & \cellsecond{0.379} & \cellsecond{1.095} & \cellsecond{0.516} & \cellsecond{3.456} & \cellsecond{0.077} & \cellsecond{2.265} \\
Ours & \cellfirst{0.343} & \cellfirst{0.830} & \cellfirst{0.433} & \cellfirst{2.022} & \cellfirst{0.065} & \cellfirst{1.899} \\
\midrule
& \multicolumn{6}{c}{\emph{Average}}\\
\midrule
PE & {2.739} & {105.973} & {1.620} & \cellthird{31.232} & \cellthird{0.311} & \cellthird{39.572} \\
Rec. LSTM & {2.292} & \cellthird{47.296} & {1.702} & {49.971} & {0.489} & {99.843} \\
Rec. Transf. & \cellthird{2.154} & {53.198} & \cellthird{1.430} & {36.123} & {0.385} & {69.977} \\
Ours Small & \cellsecond{1.084} & \cellfirst{4.461} & \cellsecond{1.077} & \cellsecond{6.016} & \cellsecond{0.140} & \cellsecond{3.590} \\
Ours & \cellfirst{1.065} & \cellsecond{4.815} & \cellfirst{0.956} & \cellfirst{4.083} & \cellfirst{0.132} & \cellfirst{3.360} \\

\bottomrule
\end{tabular}
\end{center}

\end{table}

\subsection{Animation Model Masking Strategies Ablation}
\label{ap:animation_model_masking_strategies_ablation}

\change{In Tab.~\ref{table:animation_tennis_masking_ablation}, we ablate the contribution of the animation model training masking strategies (see Sec.~\ref{ap:training_details_animation}) on the Tennis dataset. We group the masking strategies in four groups of semantically-related strategies: ``Random'' (i + ii) which masks random sequence elements, ``Block'' (iii + iv) which masks blocks of contiguous timesteps, ``Last'' (v) which masks the last timesteps of the sequence, ``Opponent'' (vi) that masks a randomly chosen set of properties in the whole sequence.}

\change{The use of only the ``Random'' masking strategy results in the worst performance as the model only learns to interpolate between adjacent values and fails on tasks requiring long predictions. Adding the ``Block'' strategy enables the model to learn how to interpolate between values separated by larger gaps, producing good results in the sequence completion task. The addition of the ``Last'' training strategy enables the model to learn how to produce future states from a given initial one, improving the results on both video prediction tasks. The ``Opponent'' masking strategy enables the model to recover properties that are missing in the whole sequence, enabling best performance on the opponent modeling task. Finally, combining all masking strategies enables the model to jointly learn these capabilities, producing the best results.}

\begin{table}[t]
\caption{\change{Ablation on the Tennis dataset of different training masking strategies of Sec.~\ref{ap:training_details_animation}: ``Random'' (i + ii), ``Block'' (iii + iv), ``Last'' (v), ``Opponent'' (vi). Position and Joints 3D in meters, Root angle in axis-angle representation.}}

\label{table:animation_tennis_masking_ablation}

\begin{center}

\footnotesize
\begin{tabular}{lcccccc}
\toprule
\multicolumn{1}{c}{}  & \multicolumn{2}{c}{\emph{Position}} & \multicolumn{2}{c}{\emph{Root angle}} & \multicolumn{2}{c}{\emph{Joints 3D}} \\
 & L2$\downarrow$  & FD$\downarrow$ & L2$\downarrow$  & FD$\downarrow$ &  L2$\downarrow$  & FD$\downarrow$ \\

\midrule
& \multicolumn{6}{c}{\emph{Action conditioned video prediction}}\\
\midrule
Random & {3.350} & {78.825} & {1.703} & {5.455} & {0.241} & {6.242} \\
Random + Block & {1.814} & {1.906} & {1.458} & {1.069} & {0.203} & {2.033} \\
Random + Last & \cellsecond{1.335} & \cellfirst{1.005} & \cellthird{1.241} & \cellthird{0.783} & \cellthird{0.186} & \cellthird{1.764} \\
Random + Opponent & \cellthird{1.355} & \cellthird{1.279} & \cellsecond{1.225} & \cellsecond{0.687} & \cellsecond{0.185} & \cellsecond{1.681} \\
All & \cellfirst{1.244} & \cellsecond{1.071} & \cellfirst{1.187} & \cellfirst{0.601} & \cellfirst{0.178} & \cellfirst{1.570} \\
\midrule
& \multicolumn{6}{c}{\emph{Unconditional video prediction}}\\
\midrule
Random & {4.583} & {227.295} & {1.663} & {5.814} & {0.287} & {12.649} \\
Random + Block & \cellthird{2.621} & \cellthird{3.995} & \cellthird{1.569} & \cellthird{0.988} & \cellthird{0.221} & {2.371} \\
Random + Last & \cellfirst{2.229} & \cellsecond{2.296} & \cellsecond{1.503} & \cellsecond{0.960} & \cellsecond{0.215} & \cellsecond{2.099} \\
Random + Opponent & {3.358} & {13.365} & {1.661} & {0.992} & {0.238} & \cellthird{2.364} \\
All & \cellsecond{2.352} & \cellfirst{2.271} & \cellfirst{1.455} & \cellfirst{0.781} & \cellfirst{0.213} & \cellfirst{1.827} \\
\midrule
& \multicolumn{6}{c}{\emph{Opponent modeling}}\\
\midrule
Random & {2.246} & {12.463} & {0.882} & {1.040} & {0.125} & {1.411} \\
Random + Block & {2.031} & {4.768} & {0.901} & \cellfirst{0.333} & \cellthird{0.123} & \cellthird{0.934} \\
Random + Last & \cellthird{1.965} & \cellthird{4.623} & \cellthird{0.861} & {0.631} & {0.125} & {1.143} \\
Random + Opponent & \cellfirst{1.486} & \cellfirst{1.191} & \cellfirst{0.823} & \cellsecond{0.359} & \cellsecond{0.115} & \cellfirst{0.755} \\
All & \cellsecond{1.578} & \cellsecond{2.243} & \cellsecond{0.832} & \cellthird{0.560} & \cellfirst{0.114} & \cellsecond{0.851} \\
\midrule
& \multicolumn{6}{c}{\emph{Sequence completion}}\\
\midrule
Random & {0.581} & {2.030} & {0.670} & {0.503} & {0.104} & {1.199} \\
Random + Block & \cellsecond{0.364} & \cellsecond{0.223} & \cellsecond{0.590} & \cellsecond{0.331} & \cellsecond{0.090} & \cellsecond{0.825} \\
Random + Last & \cellthird{0.414} & \cellthird{0.387} & \cellthird{0.614} & \cellthird{0.350} & \cellthird{0.093} & {0.904} \\
Random + Opponent & {0.494} & {0.608} & {0.659} & {0.403} & {0.096} & \cellthird{0.896} \\
All & \cellfirst{0.344} & \cellfirst{0.187} & \cellfirst{0.581} & \cellfirst{0.301} & \cellfirst{0.088} & \cellfirst{0.765} \\
\midrule
& \multicolumn{6}{c}{\emph{Average}}\\
\midrule
Random & {2.690} & {80.153} & {1.229} & {3.203} & {0.189} & {5.375} \\
Random + Block & {1.707} & \cellthird{2.723} & {1.129} & \cellthird{0.680} & {0.159} & {1.541} \\
Random + Last & \cellsecond{1.486} & \cellsecond{2.078} & \cellsecond{1.055} & {0.681} & \cellsecond{0.155} & \cellthird{1.477} \\
Random + Opponent & \cellthird{1.673} & {4.111} & \cellthird{1.092} & \cellsecond{0.610} & \cellthird{0.158} & \cellsecond{1.424} \\
All & \cellfirst{1.380} & \cellfirst{1.443} & \cellfirst{1.014} & \cellfirst{0.560} & \cellfirst{0.148} & \cellfirst{1.253} \\

\bottomrule
\end{tabular}
\end{center}
\end{table}

\subsection{Animation Model Dataset Size Ablation}
\label{ap:animation_model_dataset_size_ablation}

\change{In Tab.~\ref{table:animation_data_ablation_tennis}, we analyze the performance of the animation model as a function of the available portion of the training data on the Tennis dataset. The model performance gradually reduces as the amount of training data shrinks. When the amount of available training data falls below $60\%$ of the original dataset size, the model overfits to the training data, yielding poor performance.}

\begin{table}[t]
\caption{\change{Animation model performance as a function of the dataset size on the Tennis dataset. Position and Joints 3D in meters, Root angle in axis-angle representation.}}

\label{table:animation_data_ablation_tennis}

\begin{center}

\footnotesize
\begin{tabular}{lcccccc}
\toprule
\multicolumn{1}{c}{}  & \multicolumn{2}{c}{\emph{Position}} & \multicolumn{2}{c}{\emph{Root angle}} & \multicolumn{2}{c}{\emph{Joints 3D}}  \\
 & L2$\downarrow$  & FD$\downarrow$ & L2$\downarrow$  & FD$\downarrow$ &  L2$\downarrow$  & FD$\downarrow$ \\

\midrule
& \multicolumn{6}{c}{\emph{Action conditioned video prediction}}\\
\midrule
20\% & {4.237} & {425.180} & {1.645} & {10.917} & {0.251} & {14.293} \\
40\% & {3.723} & {244.390} & {1.543} & {6.925} & {0.236} & {9.027} \\
60\% & \cellthird{1.331} & \cellthird{1.403} & \cellthird{1.236} & \cellthird{0.954} & \cellthird{0.189} & \cellthird{1.869} \\
80\% & \cellsecond{1.283} & \cellfirst{1.024} & \cellsecond{1.207} & \cellsecond{0.714} & \cellsecond{0.183} & \cellsecond{1.607} \\
100\% & \cellfirst{1.244} & \cellsecond{1.071} & \cellfirst{1.187} & \cellfirst{0.601} & \cellfirst{0.178} & \cellfirst{1.570} \\
\midrule
& \multicolumn{6}{c}{\emph{Unconditional video prediction}}\\
\midrule
20\% & {4.391} & {445.758} & {1.736} & {12.150} & {0.260} & {15.167} \\
40\% & {4.287} & {323.800} & {1.701} & {6.922} & {0.254} & {10.196} \\
60\% & \cellthird{2.446} & \cellthird{4.059} & \cellthird{1.537} & \cellthird{1.216} & \cellthird{0.222} & \cellthird{2.257} \\
80\% & \cellsecond{2.402} & \cellsecond{2.475} & \cellsecond{1.510} & \cellsecond{0.955} & \cellsecond{0.216} & \cellsecond{1.892} \\
100\% & \cellfirst{2.352} & \cellfirst{2.271} & \cellfirst{1.455} & \cellfirst{0.781} & \cellfirst{0.213} & \cellfirst{1.827} \\
\midrule
& \multicolumn{6}{c}{\emph{Opponent modeling}}\\
\midrule
20\% & {3.829} & {321.502} & {0.993} & {2.212} & {0.130} & {2.469} \\
40\% & {3.086} & {170.115} & {0.937} & {1.771} & {0.129} & {1.979} \\
60\% & \cellthird{1.673} & \cellsecond{2.094} & \cellthird{0.837} & \cellfirst{0.416} & \cellthird{0.119} & \cellthird{0.862} \\
80\% & \cellfirst{1.550} & \cellfirst{1.606} & \cellfirst{0.817} & \cellsecond{0.450} & \cellsecond{0.114} & \cellfirst{0.749} \\
100\% & \cellsecond{1.578} & \cellthird{2.243} & \cellsecond{0.832} & \cellthird{0.560} & \cellfirst{0.114} & \cellsecond{0.851} \\
\midrule
& \multicolumn{6}{c}{\emph{Sequence completion}}\\
\midrule
20\% & {1.456} & {71.621} & {0.738} & {2.809} & {0.118} & {3.492} \\
40\% & {1.184} & {37.588} & {0.706} & {1.808} & {0.111} & {2.633} \\
60\% & \cellsecond{0.370} & \cellsecond{0.228} & \cellthird{0.608} & \cellsecond{0.291} & \cellthird{0.094} & \cellthird{0.853} \\
80\% & \cellthird{0.372} & \cellthird{0.237} & \cellsecond{0.589} & \cellfirst{0.231} & \cellsecond{0.090} & \cellsecond{0.773} \\
100\% & \cellfirst{0.344} & \cellfirst{0.187} & \cellfirst{0.581} & \cellthird{0.301} & \cellfirst{0.088} & \cellfirst{0.765} \\
\midrule
& \multicolumn{6}{c}{\emph{Average}}\\
\midrule
20\% & {3.478} & {316.015} & {1.278} & {7.022} & {0.190} & {8.855} \\
40\% & {3.070} & {193.973} & {1.222} & {4.357} & {0.183} & {5.959} \\
60\% & \cellthird{1.455} & \cellthird{1.946} & \cellthird{1.054} & \cellthird{0.719} & \cellthird{0.156} & \cellthird{1.460} \\
80\% & \cellsecond{1.402} & \cellfirst{1.335} & \cellsecond{1.031} & \cellsecond{0.588} & \cellsecond{0.151} & \cellsecond{1.255} \\
100\% & \cellfirst{1.380} & \cellsecond{1.443} & \cellfirst{1.014} & \cellfirst{0.560} & \cellfirst{0.148} & \cellfirst{1.253} \\

\bottomrule
\end{tabular}
\end{center}

\end{table}

\subsection{Alternative Samplers}
\begin{figure}
\includegraphics[width=\columnwidth]{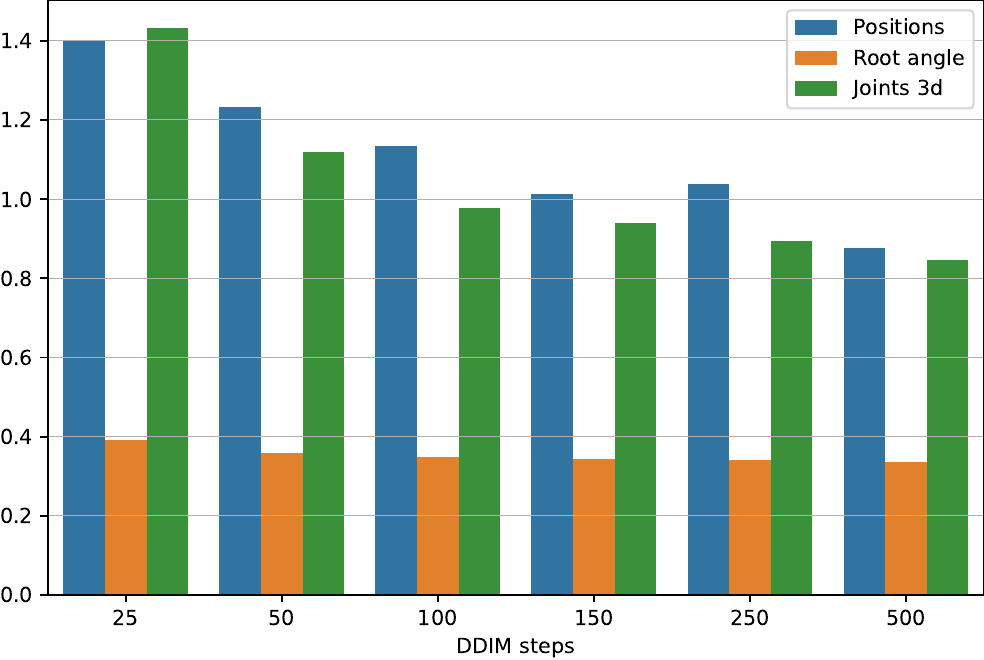}
  \caption{Evaluation results for our method on the Tennis dataset using DDIM sampler with a varying number of sampling steps.}
  \label{fig:ddim_sampling_plot}
\end{figure}

In this section, we evaluate our animation model using the DDIM \cite{song2021denoising} sampler with a varying number of timesteps and show results in Fig.~\ref{fig:ddim_sampling_plot}. The DDIM sampler produces samples with a lower number of sampling timesteps with respect to the DDPM \cite{ho2020ddpm} sampler at the cost of higher FD scores, thus providing a tradeoff between inference speed and sample quality. We note that several techniques exist that speed up diffusion model sampling \cite{salimans2022progressive,meng2022on} and that these efforts are orthogonal to our work.

\section{Discussion}

\subsection{Cost-Quality Tradeoff and Game Developers Validation}
\label{ap:cost_quality}

Building video games is an extremely expensive process. Our \change{PGM}, being a \emph{fully learnable} solution, requiring only annotated monocular videos and supporting a core set of game functions has potential applications aimed at reducing game development effort. 

This research direction is a step towards creating games and editing videos without expensive equipment, data, 3D assets, sophisticated software and manual labor of trained experts. We do not aim to surpass the quality of high-cost techniques (\$100k-\$1M, see below) that require such resources.

Evaluation of \change{PGMs} in such context necessarily needs to be performed under the light of a Pareto curve representing the tradeoff between development cost and output quality. 
We analyze three points on this curve for tennis:
\begin{itemize}
\item \emph{Our scenario: Neural video game simulation} (10k\$ cost range,
medium quality) We annotate monocular videos with granular text through a hired professional labeling team for
883\$/video-hour, totaling 13,672\$, and spend a comparable amount of compute for the remaining annotation and training our models. The model produces renderings of higher quality than state-of-the-art works operating under the same data and cost assumptions \cite{Menapace2022PlayableEnvironments}, learns a capable game AI, and is based on rapidly improving NeRF and diffusion techniques.
\item \emph{Traditional game development} (1M\$ cost range, high quality). We interviewed three game development experts with backgrounds in real-time graphics, 3D models and animation, and game development management with 45 years of combined experience. We invited the experts to discuss the recent “AO Tennis 2” game and compare it with our method. Their cost estimates for building AO Tennis 2 using existing game engines were respectively of \$100k-500k (in the US), \$600k (45-person-years in Ukraine), and of \$1M (3-person-years in the US), including software licenses, equipment and assets. They noted that ``[our model’s] game AI is very valuable and it's going to be the hardest part of developing tennis'' and that our model's game AI has the potential to be ``game changer'' for tasks such as realistically modeling the behavior of animals inserted in a game. With regards to graphics, they noted that ``[our model’s graphics] is more realistic'' and that ``[our model’s output] looks like a real video'' when not zoomed in, but commented that users may prefer a less-realistic game-like graphics because they are more accustomed to its look. The users highlighted the value of the generated animated 3D assets, remarking that high-quality 3D assets of real players are expensive. When asked about possible immediate uses of the model in game production they reported that while the model is ``impressive'' it is not yet ``mind-blowing'' when speaking about building products using our method. In particular, the model's output may present artifacts that would require correction, preventing direct use of the model in a production environment. An interviewee highlighted that the framework could be called a ``limited game engine'' and would be ``awesome'' to use to create a new type of ``promo games'' such as Superbowl or Nascar games that can be released at low cost immediately after the sport season, leveraging the captured footage. Given the significantly lower cost of our method and the rapid evolution in neural rendering and diffusion models on which our framework is based, we consider these comments an encouraging validation of this research direction.
\item \emph{Specialized CG techniques} (100k\$ cost range, high quality). Specialized character animation techniques \cite{starke2019neural,starke2020local,holden2020learned} produce high-quality animations. However, they come at higher cost. The sole requirement of motion capture data entails a professional multicamera system (1k-10k\$ per camera * 10s of cameras), motion capture software licenses (1k-10k\$/year), an HPC system with TBs of storage (>>10k\$), dedicated engineers (10k-100k\$/year), studio space, inviting professional actors or players (10-100\$/h). In addition, they do not model the complete game’s dynamics and only learn basic game AI elements, making their integration into a unified, learnable framework nontrivial.
\end{itemize}

\subsection{Choice of Hyperparameters}
\label{ap:hyperparameters_choice}

\change{The composable nature of our framework allows each object in the environment to be represented and parametrized independently from the other objects. While this offers flexibility, it introduces several hyperparameters to be set. To ease the configuration of the framework for new datasets, in the following, we summarize the main hyperparameters to be configured and the rationale on how to configure them:}
\begin{itemize}
\item \change{The set of objects to model. This is typically strongly suggested by the scene and the number of contained agents.}

\item \change{The dimension of the bounding boxes in meters for each object. This is typically known a priori for each object.}

\item \change{The structure of the kinematic tree of deformable objects. This is typically known a priori based on the method used to obtain the 3D pose estimates (SMPL for humans, internal Minecraft representation for Minecraft)}

\item \change{The type of NeRF canonical volume representation $\netcanonical$ to use for each object (see Sec.~\ref{sec:canonical_volume}). This is typically strongly suggested by the structure of the object to be modeled, i.e. 2D feature plane for objects well approximated by planes such as the tennis field, 3D feature grids for non-planar objects, and the skybox representation for the sky.}

\item \change{Dimension and number of features for the chosen representation of $\netcanonical$ (see \apref{ap:implementation_details_synthesis}). Despite the large differences between players in the Tennis and Minecraft dataset, we use the same hyperparameters, thus believe they can generalize well across datasets.
Feature planes for the tennis field and skyboxes are computationally inexpensive, thus we set a high resolution for them without tuning and assign a larger number of features to Tennis planes due to their higher level of detail.
Due to the large dimension of the Minecraft scene, we assign it a larger-resolution feature grid 128x128x128 rather than the 32x32x32 used for players.
These values can be raised in case of increased geometric complexity and level of detail in the object with respect to the showcased datasets.}

\item \change{Number of points to sample for each object category (see \apref{ap:implementation_details_synthesis}). For both Minecraft and Tennis we use 32 points for players, thus we believe the value can generalize to articulated objects in different datasets. For planar objects and skyboxes a single point must be sampled. For the Minecraft scene, we use 48 due to its larger dimension. These values can be reduced or increased based on the size and geometric complexity of the objects in the dataset of interest.}

\end{itemize}

\subsection{Ethics}
\label{ap:ethics}

The techniques described in this work fall in the category of video editing methods and could potentially be used to nefariously alter existing videos. The design of our method assumes multiple camera-calibrated observations of a single scene to be available for training, and that the desired edit is shown at least once in the training data. This provides protection against applying the method to tamper a single video, for which the quantity of data would not be sufficient and the desired edit would likely not be shown.

\end{document}